\documentclass{article}

     \PassOptionsToPackage{numbers, sort&compress}{natbib}

     \usepackage[final]{neurips_2024}

\usepackage[utf8]{inputenc} 
\usepackage[T1]{fontenc}    
\usepackage[colorlinks,linktoc=all]{hyperref}       
\usepackage[dvipsnames]{xcolor}
\hypersetup{linkcolor=MidnightBlue,citecolor=MidnightBlue,urlcolor=MidnightBlue}
\usepackage{url}            
\usepackage{booktabs}       
\usepackage{amsfonts}       
\usepackage{nicefrac}       
\usepackage{microtype}      
\usepackage{xcolor}         

\usepackage{amsmath, amssymb, amsthm, bbm, dsfont, mathrsfs}
\usepackage{xr}
\usepackage{subfiles, subcaption}
\usepackage{color, verbatim, graphicx}
\usepackage{enumitem}
\usepackage{mathtools}
\usepackage[nameinlink]{cleveref}
\creflabelformat{equation}{#2{}#1{}#3}
\crefname{equation}{eq.}{eqs.}
\Crefname{equation}{Eq.}{Eqs.}
\usepackage{multirow,multicol}

\theoremstyle{definition}

\renewcommand{\epsilon}{\varepsilon}

\newcommand{\given}{\,|\,}

\newcommand{\norm}[1]{\left\lVert#1\right\rVert}

\newcommand{\C}{\mathcal{C}}
\newcommand{\D}{\mathscr{D}}

\newcommand{\N}{\mathcal{N}}
\renewcommand{\O}{\mathcal{O}}
\newcommand{\R}{\mathcal{R}}
\newcommand{\Q}{\mathcal{Q}}

\newcommand{\KL}{\text{KL}}

\newcommand{\reals}{\mathbb{R}}

\newtheorem{theorem}{Theorem}[section]

\newtheorem{lemma}[theorem]{Lemma}

\usepackage{thmtools}

\definecolor{shade}{rgb}{0.9,0.9,0.9}

\DeclareMathOperator*{\argmin}{argmin}
\renewcommand{\R}{\mathbb{R}}

\title{EigenVI:\ score-based variational inference with orthogonal function expansions}

\author{%
    Diana Cai \\
    Flatiron Institute \\
    \texttt{dcai@flatironinstitute.org} \\
  \And
    Chirag Modi \\
    Flatiron Institute \\
  \texttt{cmodi@flatironinstitute.org} \\
  \And
    Charles C.\ Margossian \\
    Flatiron Institute \\
  \texttt{cmargossian@flatironinstitute.org} \\
  \And
    Robert M.\ Gower \\
    Flatiron Institute \\
  \texttt{rgower@flatironinstitute.org} \\
 \AND
    David M.\ Blei \\
    Columbia University \\
  \texttt{david.blei@columbia.edu} \\
  \And
    Lawrence K.\ Saul \\
    Flatiron Institute \\
  \texttt{lsaul@flatironinstitute.org} \\
}

\begin{document}

\maketitle

\begin{abstract}
We develop EigenVI, an eigenvalue-based approach for black-box variational inference (BBVI). EigenVI constructs its variational approximations from orthogonal function expansions. For distributions over~$\mathbb{R}^D$, the lowest order term in these expansions provides a Gaussian variational approximation, while higher-order terms provide a systematic way to model non-Gaussianity. These approximations are flexible enough to model complex distributions (multimodal, asymmetric), but they are simple enough that one can calculate their low-order moments and draw samples from them. EigenVI can also model other types of random variables (e.g., nonnegative, bounded) by constructing variational approximations from different families of orthogonal functions. Within these families, EigenVI computes the variational approximation that best matches the score function of the target distribution by minimizing a stochastic estimate of the Fisher divergence. Notably, this optimization reduces to solving a minimum eigenvalue problem, so that EigenVI effectively sidesteps the iterative gradient-based optimizations that are required for many other BBVI algorithms. (Gradient-based methods can be sensitive to learning rates, termination criteria, and other tunable hyperparameters.) We use EigenVI to approximate a variety of target distributions, including a benchmark suite of Bayesian models from \texttt{posteriordb}. On these distributions, we find that EigenVI is more accurate than existing methods for Gaussian BBVI.

\end{abstract}

\section{Introduction}



Probabilistic modeling is a cornerstone of modern data analysis,
uncertainty quantification, and decision making. A key challenge of
probabilistic inference is computing a target distribution of
interest; for instance, in Bayesian modeling, the goal is to compute a
posterior distribution, which is often intractable. Variational
inference (VI) \citep{jordan1999vi,wainwright2008graphical,blei2017vi}
is a popular method for scalable probabilistic inference that has worked across a range of applications. The idea behind VI is to approximate the target distribution by the closest member of some tractable family.

One major focus of research is to develop \textit{black-box} algorithms for variational inference
\citep{ranganath2014black,kingma2013auto,titsias2014doubly,kucukelbir2017automatic,  locatello2018boosting,giordano2023black,wang2022dual,modi2023,cai2024}.
Algorithms for black-box variational inference (BBVI) can be used to approximate any target distribution that is differentiable and computable up to some multiplicative (normalizing) constant; as such, they are extremely flexible. These algorithms have been widely implemented in popular probabilistic programming languages, and they are part of the modern toolbox for practitioners in computational statistics and data analysis~\citep{salvatier2016probabilistic,carpenter2017stan,ge2018turing,bingham2019pyro,abril2023pymc}.

Traditionally, the variational approximations in BBVI
are optimized by minimizing the Kullback-Leibler (KL)
divergence between the variational family and the target
(equivalently, maximizing the ELBO).
This strategy is powerful and
scalable, but it relies on stochastic gradient descent (SGD), which can be
difficult to tune
\citep{dhaka2020robust,dhaka2021challenges,zhang2022pathfinder}.
These difficulties can be acute even for Gaussian variational approximations~\citep{ranganath2014black,kucukelbir2017automatic}, particularly if these approximations employ full covariance matrices.

More recently, researchers have proposed algorithms for Gaussian BBVI that
do not require the use of SGD \citep{modi2023,cai2024}. Instead of minimizing the KL divergence, these methods aim to match the \emph{scores}, or the gradients of the log densities,
between the variational distribution and the target density. These methods exploit the special form of Gaussian distributions to derive closed-form proximal point
updates for score-matching. These updates are as inexpensive as SGD, but not as brittle. They show that score-based BBVI can be applied in an elegant way to Gaussian variational families.

In this paper, we show that score-based BBVI also yields simple, closed-form updates for a much broader family of variational approximations. Specifically, we
propose a new class of variational families constructed from
\textit{orthogonal function expansions} and inspired by solutions to the Schr\"odinger equation in quantum mechanics. These families are expressive enough to
parameterize a wide range of target distributions; at the same time, the distributions in these families are sufficiently tractable that one can calculate low-order moments and draw samples from them. In this paper, we mostly use orthogonal function expansions to construct distributions supported on $\reals^D$; in this case, the lowest-order term in the expansion is sufficient to model Gaussian behavior, while higher-order terms account for increasing amounts of non-Gaussianity. More generally, we also show how different basis sets of orthogonal functions can be used to construct variational families over other spaces.

To optimize over a variational family from this class, we minimize an
estimate of the Fisher divergence, which measures the scores of the
variational distribution against those of the target distribution. We
show that this optimization reduces to a minimum
eigenvalue problem, thus avoiding the need for gradient-based methods.
For this reason, we call our approach \emph{EigenVI}.

We study EigenVI with a variational family constructed from weighted
Hermite polynomials. We first demonstrate the expressiveness of this family on a
variety of
multimodal, asymmetric, and heavy-tailed distributions. We then use EigenVI to approximate a diverse collection of
non-Gaussian target distributions from \texttt{posteriordb}
\citep{magnusson2022posteriordb}, a benchmark suite of Bayesian
hierarchical models. On these problems, EigenVI provides more accurate posterior approximations than leading implementations of Gaussian
BBVI based on KL minimization and score-matching.

The organization of this paper is as follows.
In \Cref{sec:scorebasedVI} we introduce the variational families that arise from orthgonal function expansions, and we
show how score-matching in these families reduces to an
eigenvalue problem. In \Cref{sec:related} we review the literature related
to EigenVI. In \Cref{sec:experiments}, we evaluate EigenVI on a
variety of synthetic and real-data targets. Finally, in \Cref{sec:conclusion},
we discuss limitations and future work.

%



\section{Score-based variational inference with orthogonal function expansions}
\label{sec:scorebasedVI}

In this section we use orthogonal function expansions to develop new variational families for
approximate probabilistic inference. In \Cref{sec:orth}, we review the basic properties of these
expansions.
In \Cref{sec:eigenVI}, we introduce a score-based divergence for VI with these families;
notably, for this divergence, the optimization for VI reduces to an eigenvalue
problem. Finally in \Cref{sec:precon},
we consider how to use these variational approximations for unstandardized distributions;
in these settings we must carefully manage the trade-off between expressiveness and
computational~cost.


\subsection{Orthogonal function expansions}
\label{sec:orth}

Let $\mathcal{Z}\subseteq\mathbb{R}^D$ denote the support of the
target distribution $p$.
Suppose that there exists a
complete set of orthonormal basis functions
$\{\phi_k(z)\}_{k=1}^\infty$ on this set. By \emph{complete}, we mean
that any sufficiently well-behaved function
\mbox{$f:\mathcal{Z}\rightarrow\mathbb{R}$} can be approximated, to
arbitrary accuracy, by a particular weighted sum of these basis
functions, and by \emph{orthonormal}, we mean that the basis functions
satisfy
\begin{align}
\label{eq-orthonormal}
  \int\!\phi_k(z)\phi_{k'}(z)\, dz =
  \left\{
  \begin{array}{rr} 1 & \mbox{if $k=k'$}, \\
    0 & \mbox{otherwise,}\end{array}\right.
\end{align}
where the integral is over $\mathcal{Z}$.
Define the
$K^\text{th}$-order variational family $\mathcal{Q}_K$ to be the set containing
all distributions of the form
\begin{align}
  q(z) =  \left(\sum_{k=1}^{K} \alpha_k \phi_k(z)\right)^2\quad\mbox{where}\quad\sum_{k=1}^{K} \alpha_k^2=1,
\label{eq:OF-1}
\end{align}
and where $\alpha_k\in\R$ for $k=1,\ldots, K$ are the parameters of the family $\mathcal{Q}_K.$
In words,  $\mathcal{Q}_K$ contains all distributions that can be obtained by
taking weighted sums of the first $K$ basis functions and then \textit{squaring} the result.

\Cref{eq:OF-1} involves a squaring operation, a sum-of-squares
constraint, and a weighted sum. The squaring operation ensures that
the density functions in $\mathcal{Q}_K$ are nonnegative (i.e., with
$q(z)\!\geq\!0$ for all $z\in\mathcal{Z}$), while the sum-of-squares constraint
ensures that they are normalized:
\begin{align}
  \int\! q(z)\, dz\, =\, \int\! \left(\sum_{k=1}^{K} \alpha_k \phi_k(z)\right)^2\!\! dz\, =\,
  \int\! \sum_{k,k'=1}^{K} \alpha_k \alpha_{k'} \phi_k(z)\phi_{k'}(z)\, dz\, =\, \sum_{k=1}^{K} \alpha_k^2 = 1.
\end{align}

The weighted sum in \Cref{eq:OF-1} bears a superficial similarity
to a mixture model, but note that neither the basis
functions~$\phi_k(z)$ nor the weights~$\alpha_k$ in \Cref{eq:OF-1}
are constrained to be nonnegative. Distributions of this form arise
naturally in physics from the quantum-mechanical \emph{wave functions}
that satisfy Schr\"odinger's equation \citep{griffiths2018introduction}.
In that setting, though, it is
typical to consider complex-valued weights and basis functions,
whereas here we only consider real-valued ones.

The simplest examples of orthogonal function expansions arise in one
dimension. For example, functions on the interval $[-1,1]$ can be
represented as weighted sums of Legendre polynomials, while functions
on the unit circle can be represented by Fourier series of sines and
cosines; see \Cref{tab:onedim}. Distributions on unbounded
intervals can also be represented in this way. On the real line, for
example, we may consider approximations of the form in
\Cref{eq:OF-1} where
\begin{equation}
\phi_{k+1}(z) = \left(\sqrt{2\pi}k!\right)^{-\frac{1}{2}}\left(e^{-\frac{1}{2}z^2}\right)^{\frac{1}{2}}\,\text{H}_{k}(z),
\label{eq:phi-hermite}
\end{equation}
and $\text{H}_k(z)$ are the \emph{probabilist's} Hermite polynomials given by
\begin{equation}
\text{H}_k(z) = (-1)^k e^{\frac{z^2}{2}} \frac{d^k}{dz^k}\left[e^{-\frac{z^2}{2}}\right].
\label{eq:hermite}
\end{equation}
Note how the lowest-order basis function $\phi_1(z)$ in this family gives rise
(upon squaring) to a Gaussian distribution with zero mean and unit variance.

\begin{figure}[t]
\centering
\begin{subfigure}[b]{0.325\linewidth}
    \centering
    \includegraphics[scale=0.31]{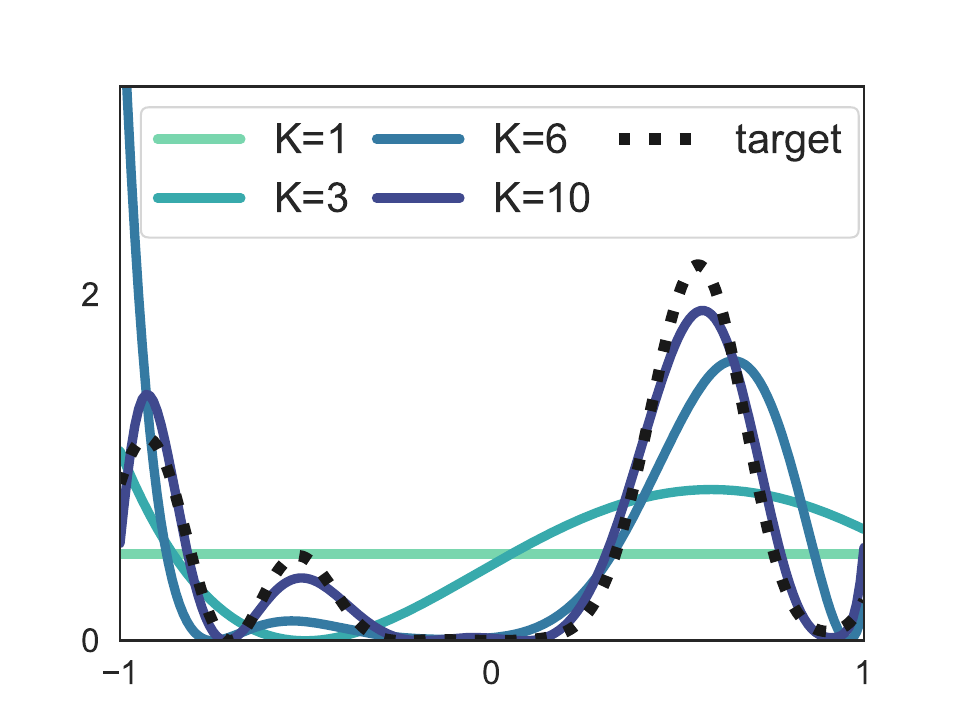}
    \caption{Legendre polynomial expansion}
\end{subfigure}
\begin{subfigure}[b]{0.325\linewidth}
    \centering
    \includegraphics[scale=0.31]{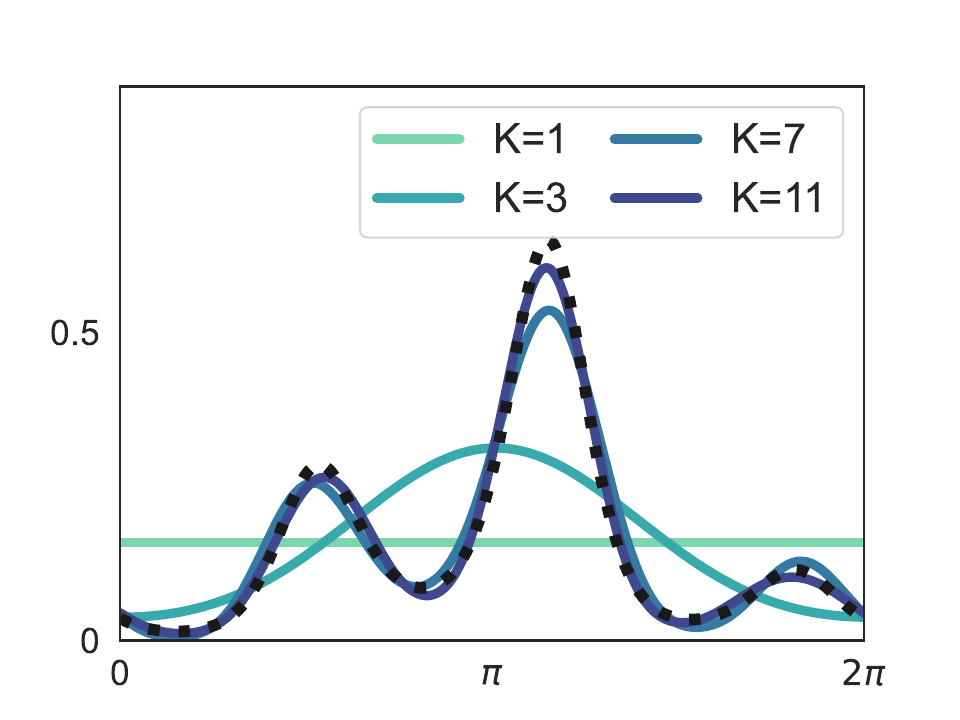}
    \caption{Fourier series expansion}
\end{subfigure}
\begin{subfigure}[b]{0.325\linewidth}
    \centering
    \includegraphics[scale=0.31]{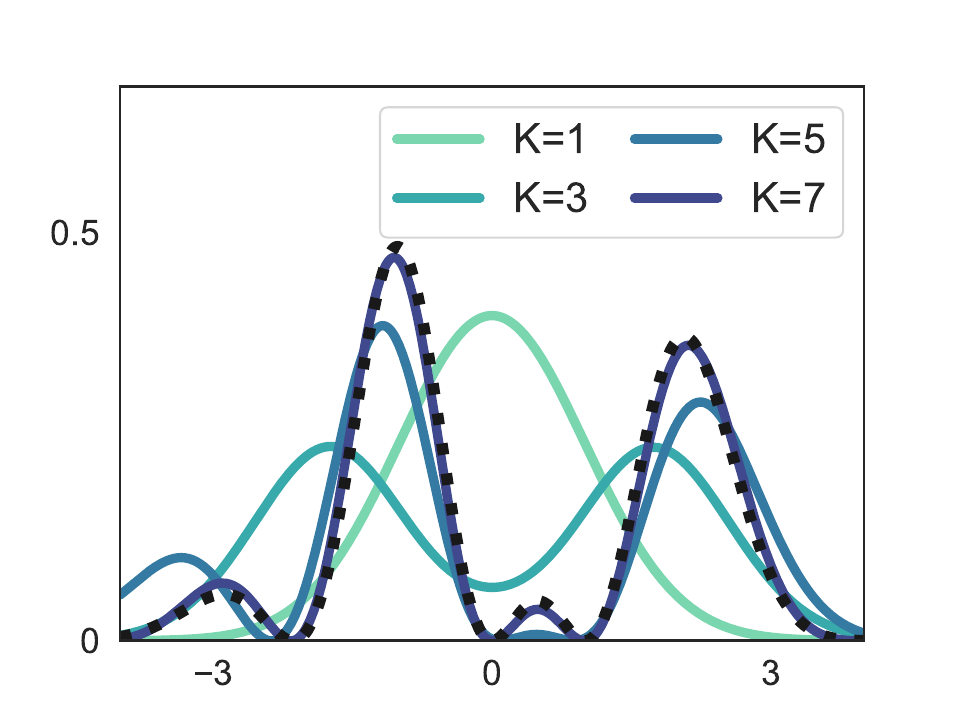}
    \caption{Hermite polynomial expansion}
    \label{fig:onedim:c}
\end{subfigure}
\caption{Target probability distributions (black dashed curves) on the interval $[-1,1]$ (left), the unit circle (middle), and the
    real line (right), and their approximations by orthogonal function expansions
    from different families and of different orders; see \Cref{eq:OF-1} and \Cref{tab:onedim}.}
\vspace{-15pt}
\label{fig:onedim}
\end{figure}

\Cref{fig:onedim} shows how various multimodal distributions with
one-dimensional support can be approximated by computing weighted sums
of basis functions and squaring their result. We emphasize that
\textit{the more basis functions in the sum, the better the
  approximation}.


\begin{table*}[t]
\small
    \centering
    \caption{Examples of orthogonal function expansions in one dimension. The basis functions in the
        table are not normalized, but they can be rescaled so that their squares integrate to one.
    \vspace{5pt}
    }

    \label{tab:onedim}
    \begin{tabular}{lll}
    \toprule
    \textbf{support} & \textbf{orthogonal family} & \textbf{basis functions} $\phi_k(\cdot)$  \\ [0.3ex]
    \midrule
    $z\in[-1,1]$ & Legendre polynomials & $\{1,\, z,\, 3z^2\!-\!1,\, 5z^3\!-\!3z, \ldots\}$ \\ [0.3ex]
    $z=e^{i\theta}\!\in S^1$\hspace{2ex} & Fourier basis & $\{1,\cos\theta,\sin\theta,\cos 2\theta,\sin 2\theta,\ldots\}$ \\[0.3ex]
    $z\in[0,\infty)$ & weighted Laguerre polynomials &    $e^{-\frac{z}{2}}\{1,1\!-\!z,z^2\!-\!4z\!+\!2,\ldots\}$\\ [-0.3ex]
    $z\in\mathbb{R}$ & weighted Hermite polynomials & $e^{-\frac{z^2}{4}}\{1,z,(z^2\!-\!1),(z^3\!-\!3z),\ldots\}$\\ [0.5ex]
    \bottomrule
    \end{tabular}
    \vspace{-10pt}
\end{table*}

Orthogonal function expansions in one dimension are also important because their Cartesian products can be used to generate orthogonal function expansions in higher dimensions. For example, we can approximate distributions over (say) $\mathbb{R}^3$ by
\begin{equation}
q(z_1,z_2,z_3) =  \left(\sum_{i=1}^{K_1}\sum_{j=1}^{K_2}\sum_{k=1}^{K_3} \beta_{ijk}\, \phi_i(z_1)\phi_j(z_2)\phi_k(z_3)\right)^2\quad\mbox{where}\quad \sum_{ijk}\beta^2_{ijk} = 1,
\label{eq:3d}
\end{equation}
where $\beta_{ijk}\in \R$ now parametrize the family.
Note that there are a total $K_1 K_2 K_3$ parameters in the above expansion, so that this method of
Cartesian products does not scale well to high dimensions if multiple basis functions are used per
dimension.
Note that the same strategy can also be used for random variables of
mixed type: for example, from \Cref{tab:onedim}, we can create a variational family of distributions over $\mathbb{R}\!\times\![-1,1]\!\times\![0,\infty)$ from the Cartesian product of orthogonal function expansions involving Hermite, Legendre, and Laguerre polynomials.

As shown in \Cref{fig:onedim}, the approximating distributions
from $K^\text{th}$-order expansions can model the presence of multiple
modes as well as many types of asymmetry, and this expressiveness also
extends to higher dimensions. Nevertheless, it remains tractable to
sample from these distributions and even to calculate (analytically)
their low-order moments, as we show in \Cref{app:sampling,app:moments}.

For concreteness, consider the distribution over $\mathbb{R}^3$ in
\Cref{eq:3d}. The marginal distribution $q(z_1)$ is
\begin{align}
q(z_1) = \int\! q(z_1,z_2,z_3)\,dz_2\, dz_3 = \sum_{ii'} \left[\sum_{jk} \beta_{ijk}\beta_{i'jk}\right] \phi_i(z_1)\phi_{i'}(z_1),
\label{eq:marginal}
\end{align}
and from this expression, moments such as $\mathbb{E}[z_1]$
and
$\text{Var}[z_1]$
can be calculated by evaluating integrals involving
the elementary functions in \Cref{tab:onedim}.
(In practice,
these integrals are further simplified by recursion relations that
relate basis functions of different orders;
we demonstrate how to compute the first two moments for
the normalized Hermite family in
\Cref{eq:mom1ij,eq:mom2ij}.)

To generate samples $\{z^{(t)}\}$, each dimension is sampled as follows:
we draw
$z_1^{(t)} \sim q(z_1)$ by computing the cumulative distribution function (CDF)
of this marginal distribution and then numerically inverting this CDF.
Finally, extending these ideas, we can calculate higher-order moments
and obtain joint samples via the nested draws
\begin{align}
\label{eq:sample}
    z_1^{(t)} \sim q(z_1),\quad z_2^{(t)} \sim q(z_2 \given z_1),\quad
    z_3^{(t)} \sim q(z_3  \given z_1,z_2).
\end{align}
The overall complexity of these procedures scales no worse than
quadratically in the number of basis functions in the expansion. These
extensions are discussed further in \Cref{app:sampling,app:moments}.


\subsection{EigenVI}
\label{sec:eigenVI}

In variational inference, we posit a parameterized family of
approximating distributions and then compute the particular approximation
in this family that is closest to a target distribution of interest.
\Cref{eq:OF-1}
constructs a variational family $\Q_K$ from the orthogonal functions $\{\phi_k(z)\}_{k=1}^K$ whose variational parameters are the weights
$\{\alpha_k\}_{k=1}^{K}$. We now derive \textit{EigenVI}, a method to find
$q \in \Q_K$ that is close to the target distribution $p(z)$.

We first define the measure of closeness that we will minimize.
EigenVI measures the quality of an approximate density by the
\textit{Fisher divergence} \citep{hyvarinen2005estimation},
\begin{align}
\label{eq-divergence}
    \D(q,p) = \int \norm{\nabla \log q(z) - \nabla\log p(z)}^2 q(z) dz,
\end{align}
where $\nabla \log q(z)$ and $\nabla \log p(z)$ are the score functions of the
variational approximation and target, respectively. Suppose that $q$ and $p$ have the same support; then the Fisher divergence vanishes if and only if the scores of $q$ and $p$ are
everywhere equal.

Though $p$ is, by assumption, intractable to compute, in many applications it is
possible to efficiently compute the score $\nabla \log p$ at any point
$z\in\mathcal{Z}$. For example, in Bayesian models the score of the target
posterior is equal to the gradient of the log joint. This observation is the main
motivation for score-based methods in probabilistic
modeling~\citep{liu2016stein,yu2023semiimplicit,modi2023,cai2024}.

Here we seek the $q\!\in\!\Q_K$ that minimizes $\D(q,p)$. But now a
challenge arises: it is generally difficult to evaluate the integral
for $\D(q,p)$ in \Cref{eq-divergence}, let alone to minimize it as a
function of~$q$.
While it is possible to sample from the distribution $q$,
it is not straightforward to simultaneously sample from $q$ and optimize over the variational parameters $\{\alpha_k\}_{k=1}^K$ in terms of which it is defined.
Instead, we construct an unbiased estimator of
$\D(q,p)$ by importance sampling, which also decouples the sampling distribution from the optimization.
Let $\{z^1,z^2,\ldots z^B\}$ denote a
batch of $B$ samples drawn from some proposal distribution $\pi$ on
$\mathcal{Z}$. From these samples we can form the unbiased estimator
\begin{align}
  \label{eq-empirical-divergence}
  \widehat\D_{\pi}(q, p) =
  \sum_{b=1}^B \frac{q(z^b)}{\pi(z^b)} \, \big\|\nabla \log q(z^b) - \nabla\log p(z^b)\big\|^2.
\end{align}
This estimator should be accurate for appropriately broad proposal
distributions and for sufficiently large batch sizes. We can therefore
attempt to minimize \Cref{eq-empirical-divergence} in place of
\Cref{eq-divergence}.

Now we show that the minimization of \Cref{eq-empirical-divergence} over
$q\in\Q_K$ simplifies to a minimum eigenvalue problem for the weights
$\{\alpha_k\}_{k=1}^{K}$.
To obtain the eigenvalue problem, we substitute the
orthogonal function expansion in \Cref{eq:OF-1}
into \Cref{eq-empirical-divergence} for the unbiased estimator of
$\D(q,p)$. As an intermediate step, we differentiate
\Cref{eq:OF-1} to obtain the scores
\begin{equation}
\label{eq:scores}
\nabla\log q(z^b) = \frac{2\sum_k \alpha_k \nabla \phi_k(z^b)}{\sum_k \alpha_k \phi_k(z^b)}.
\end{equation}
Further substitution of the scores provides the key result behind our
approach: the unbiased estimator in
\Cref{eq-empirical-divergence} is a simple quadratic form in the
weights $\alpha := [\alpha_1,\ldots,\alpha_K]^\top$ of the orthogonal function expansion,
\begin{equation}
\widehat\D_{\pi}(q,p) =
\alpha^\top\! M\alpha,
\label{eq:quadform}
\end{equation}
where the coefficients of the quadratic form are given by
\begin{equation} \label{eq:M}
  M_{jk} = \sum_{b=1}^B\frac{1}{\pi(z^b)}
    \left[2\nabla \phi_j(z^b) - \phi_j(z^b)\nabla\log p(z^b)\right]\cdot\left[2\nabla \phi_k(z^b) -
    \phi_k(z^b)\nabla\log p(z^b)\right].
\end{equation}
Note that the elements of the $K\!\times\! K$ symmetric matrix $M$ capture all of the dependence on
the batch of samples $\{z^b\}_{b=1}^B$, the scores of $p$ and $q$ at these samples,
and
the choice of the family of orthogonal functions.
Next we minimize the quadratic
form in \Cref{eq:quadform} subject to the sum-of-squares
constraint $\sum_k \alpha_k^2 = 1$ in \Cref{eq:OF-1}.
In this way we obtain the eigenvalue problem~\citep{courant1924methoden}
\begin{align}
  \label{eq:eigenVI-solution}
\min_{q\in\Q_K}\left[\widehat{\D}_{\pi}(q,p)\right] =
  \min_{\|\alpha\|=1}
  \left[\alpha^\top M\alpha\right] =:
  \lambda_{\text{min}}(M),
\end{align}
where $\lambda_{\text{min}}(M)$ is the minimal eigenvalue of $M$,  and the
optimal weights are given (up to an arbitrary sign) by its
corresponding eigenvector; see \Cref{app:eigen} for a proof.
EigenVI solves \Cref{eq:eigenVI-solution}.

We note that the eigenvalue problem in EigenVI arises from the curious alignment of three particular choices---namely, (i) the choice of variational family
(based on orthogonal function expansions), (ii) the choice of
divergence (based on score-matching), {and (iii) the choice of estimator for the divergence
(based on importance sampling)}.
The simplicity of this eigenvalue problem stands
in contrast to the many heuristics of gradient-based optimizations---involving learning
rates, terminating criteria, and perhaps other algorithmic
hyperparameters---that are typically required for ELBO-based
BBVI~\citep{dhaka2020robust,dhaka2021challenges}.
But EigenVI is also not entirely free of heuristics; to compute the estimator in
\Cref{eq-empirical-divergence} we must also specify the proposal distribution $\pi$ and the number
of samples $B$; see \Cref{sec:discussion:eigenvi} for a discussion.

The size of the eigenvalue problem in EigenVI is equal to the
number of basis functions $K$ in the orthogonal function expansion of \Cref{eq:OF-1}. The
eigenvalue problem also generalizes to orthogonal function expansions
that are formed from Cartesian products of one-dimensional families,
but in this case, if multiple basis functions are used per dimension,
then the overall basis size grows exponentially in the dimensionality.
Thus, for example, the eigenvalue problem would be of size
$K_1 K_2 K_3$ for the approximation in \Cref{eq:3d}, as can be
seen by ``flattening'' the tensor of weights $\beta$ in \Cref{eq:3d}
into the vector of weights $\alpha=\textbf{vec}(\beta)$ in \Cref{eq:OF-1}.
Finally, we note that EigenVI only needs to compute the minimal eigenvector of $M$ in \Cref{eq:eigenVI-solution}, and therefore it can benefit from specialized routines that are much less expensive than a full diagonalization.

\subsection{EigenVI in $\reals^D$:\ the Hermite family and standardization}
\label{sec:precon}

We now discuss the specific case of EigenVI for $\mathcal{Z}\!=\!\reals^D$ with the Hermite-based variational family in \Cref{eq:phi-hermite}.
For this case, we propose a transformation of the domain that serves to precondition or \textit{standardize}
the target distribution before applying EigenVI.
While this standardization is not required to use EigenVI,
it  helps to reduce the number of basis functions needed to approximate the target,
leading to a more computationally efficient procedure.
It also suggests natural default choices for the
proposal distribution $\pi$ in \Cref{eq-empirical-divergence}.

Recall that the eigenvalue problem grows linearly in size with the number of basis functions. Before applying EigenVI, our goal is therefore to transform the domain in a way that reduces the number of basis functions needed for a good approximation.
To meet this goal for distributions over $\reals^D$, we observe that the lowest-order basis function of the Hermite family in \Cref{eq:phi-hermite} yields (upon squaring) a standard multivariate Gaussian, with zero mean and unit covariance.
Intuitively, we might expect the approximation of EigenVI to require fewer basis functions if the statistics of the target distribution nearly match those of this lowest-order basis function. The goal of standardization is to achieve this match, to whatever extent possible, by a suitable transformation of the underlying domain. Having done so, EigenVI in $\reals^D$ can then be viewed as
a systematic framework to model non-Gaussian effects via a small number of higher-order terms in its orthogonal function expansion.

Concretely, we consider a linear transformation  of the domain:
\begin{align}
    \tilde{z} =
    \Sigma^{-\frac{1}{2}}(z\!-\!\mu),
\end{align}
where $\mu$ and $\Sigma$ are  estimates of the mean and covariance
obtained
from some other algorithm
(e.g., a Laplace approximation,  Gaussian variational inference,  Monte Carlo, or domain-specific
knowledge).
We then apply the EigenVI to fit a $K$th-order variational approximation $\tilde q(\tilde z)$
to the target distribution $\tilde{p}(\tilde{z})$ that is induced by this transformation;
afterwards, we reverse the change-of-variables to obtain the final approximation to $p(z)$, i.e.,
\begin{align}
q(z) = \tilde{q}(\tilde{z})|\Sigma|^{-1/2}.
\end{align}

\begin{figure}
\centering
\begin{subfigure}[b]{0.40\linewidth}
    \centering
    \includegraphics[scale=0.31]{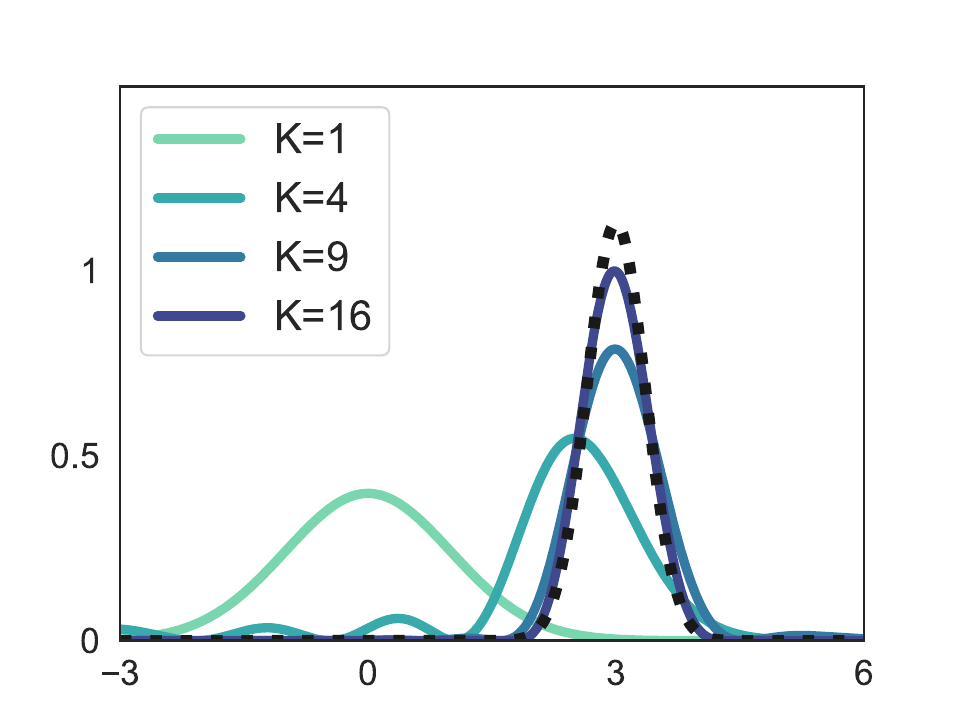}
    \caption{Gaussian target, mean $3$ and variance $\frac{1}{8}$}
\end{subfigure}
\begin{subfigure}[b]{0.40\linewidth}
    \centering
    \includegraphics[scale=0.31]{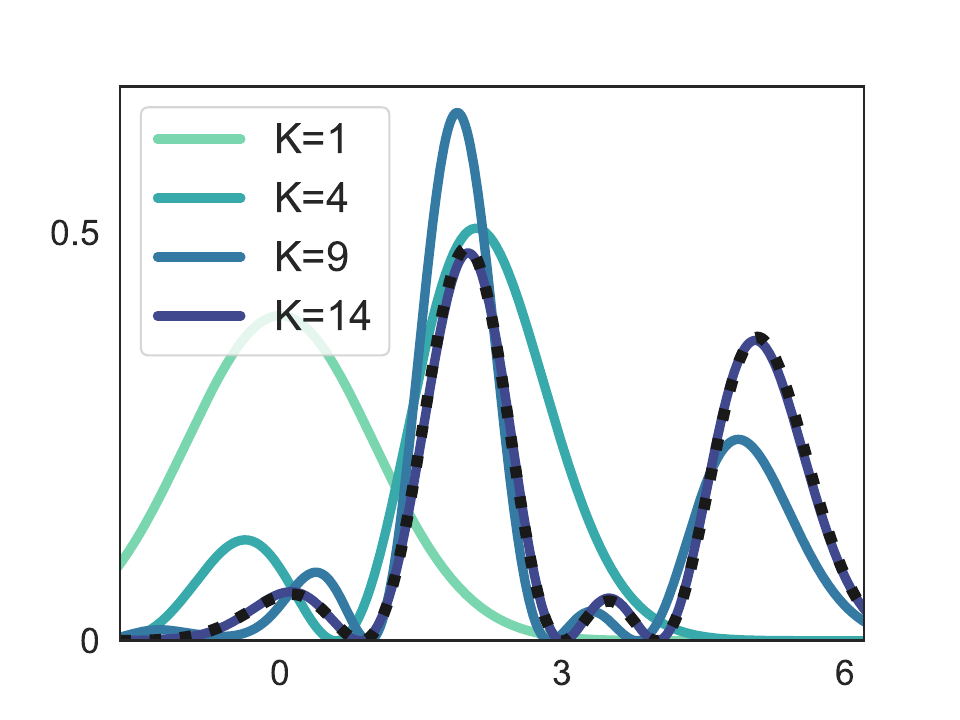}
\caption{Mixture target (translation of \Cref{fig:onedim:c})}
\end{subfigure}
    \caption{Higher-order expansions may be required to approximate target distributions
(black)
    that are not standardized. \textit{Left:} approximation of a non-standardized Gaussian.
    \textit{Right:} approximation of the mixture distribution
    in \Cref{fig:onedim} after translating its largest modes away from the origin.
}
\label{fig:standardize}
\vspace{-13pt}
\end{figure}

\Cref{fig:standardize} shows why it is more difficult to approximate distributions that are
badly centered or poorly scaled. The left panel shows the effect of translating a standard Gaussian
\textit{away} from the origin and \textit{shrinking} its variance; note how a comparable
approximation to the uncentered Gaussian now requires a 16th-order expansion.
On the other hand, after standardization, the target
can be perfectly fitted by the base distribution
in the orthogonal family of reweighted Hermite polynomials.
The right panel shows
the similar effect of translating the mixture distribution in \Cref{fig:onedim} (right panel);
comparing these panels, we see that twice as many basis functions ($K\!=\!14$ versus $K\!=\!7$)
are required to provide a comparable fit of the uncentered mixture.

Finally, we note another benefit of standardizing the target before fitting EigenVI; when the target has nearly zero mean and unit covariance, it becomes simpler to identify natural choices for the proposal distribution $\pi$.
Intuitively, in this case, we want a proposal distribution that has the same mean but heavier tails than a standard Gaussian.
In our experiments, we use two types of centered proposal distributions---uniform and isotropic Gaussian---whose variances are greater than one.





\section{Related work}
\label{sec:related}
\vspace{-4pt}

Several recent works have considered BBVI
methods based on score-matching.
These methods take a particularly simple form for
Gaussian variational families~\citep{modi2023,cai2024}.
The Fisher divergence \citep{hyvarinen2005estimation}
has been previously studied as a divergence
for variational inference \citep{yang2019variational}.
\citet{yu2023semiimplicit} propose minimizing a Fisher divergence
for semi-implicit (non-Gaussian) variational families; the divergence is minimized
using gradient-based optimization.
In another line of work,
\citet{zhang2018variational}
consider variational families of energy-based models
and derive a closed-form solution to minimize the Fisher divergence in this setting.

{More generally, there have many studies of VI with non-Gaussian variational families.
One common extension is to consider families of mixture models
\citep{guo2016boosting,miller2017variational,gershman2012nonparametric};
these are typically optimized via ELBO maximization.
}
BBVI algorithms have also been derived for
more expressive variational families of
energy-based models \citep{zhu1998filters,lecun2006tutorial,kim2016deep,dai2019kernel,lawson2019energy,zoltowski2021slice}
and normalizing flows
\citep{rezende2015variational,kingma2016improved,louizos2017multiplicative,berg2018sylvester,kobyzev2020normalizing,papamakarios2021normalizing}.
However the performance of these models, especially the normalizing flows, is often sensitive to the hyperparameters of the flow architecture and optimizer, as well as the parameters of the base distribution~\citep{dhaka2021challenges, agrawal2020advances}.
Other aspects of these variational approximations are also less straightforward; for example, one cannot compute their low-order moments, and one cannot easily evaluate or draw samples from the densities of energy-based models.

The variational approximation in EigenVI is based on the idea of squaring a weighted sum of basis functions. Probability distributions of this form arise most famously in quantum mechanics \citep{griffiths2018introduction}.
This idea has also been used to model distributions in machine learning, though not quite in the way proposed here.
\citet{novikov2021tensor} propose a tensor train-based model for density estimation, but they do not
consider orthogonal basis sets.
Similarly, \citet{loconte2024subtractive} obtain distributions by squaring a mixture model with
negative weights, and they study this model in conjunction with probabilistic circuits.
By contrast in this work, we consider this idea in the context of variational inference,
and we focus specifically on the use of orthogonal function expansions, which have many simplifying
properties;
additionally, the specific objective we optimize leads to a minimum eigenvalue problem.




\vspace{-5pt}
\section{Experiments}
\label{sec:experiments}
\vspace{-3pt}

We evaluate EigenVI on 9 synthetic targets and 8 real data targets.
In these experiments we use the orthogonal family induced by
\emph{normalized Hermite polynomials} (see \Cref{tab:onedim}),
whose lowest-order expansion is  Gaussian.
Thus, this variational family can model non-Gaussian behavior with the higher-order functions in its basis.
We first study 2D synthetic targets and use them to demonstrate the expressiveness of
these higher-order expansions.
Next, we experiment with target distributions where we systematically vary the tail heaviness and amount of skew.
Finally, we apply EigenVI to a set of hierarchical Bayesian models from real-world
applications and benchmark its performance against other Gaussian BBVI algorithms.

\subsection{2D synthetic targets}

\begin{figure*}[t]
    \centering
    \includegraphics[scale=0.44]{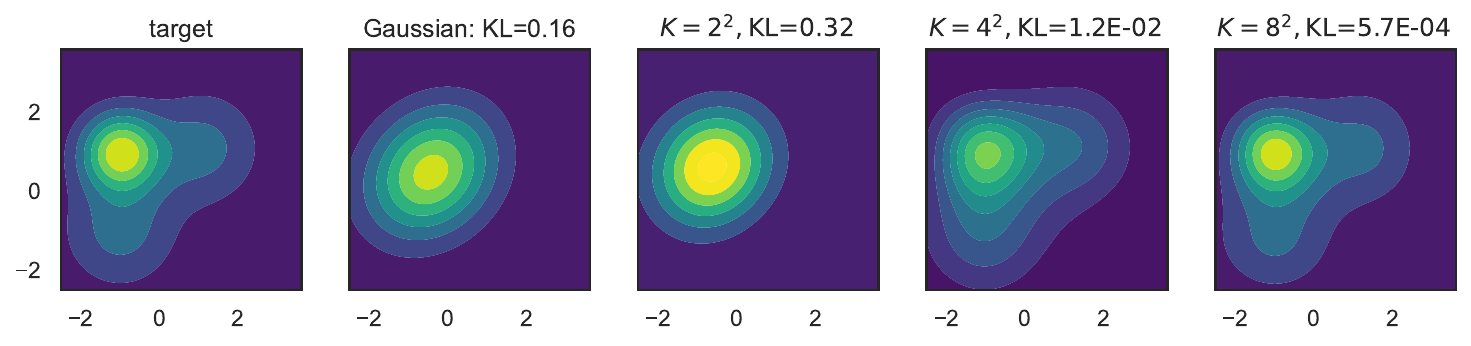}
    \includegraphics[scale=0.44]{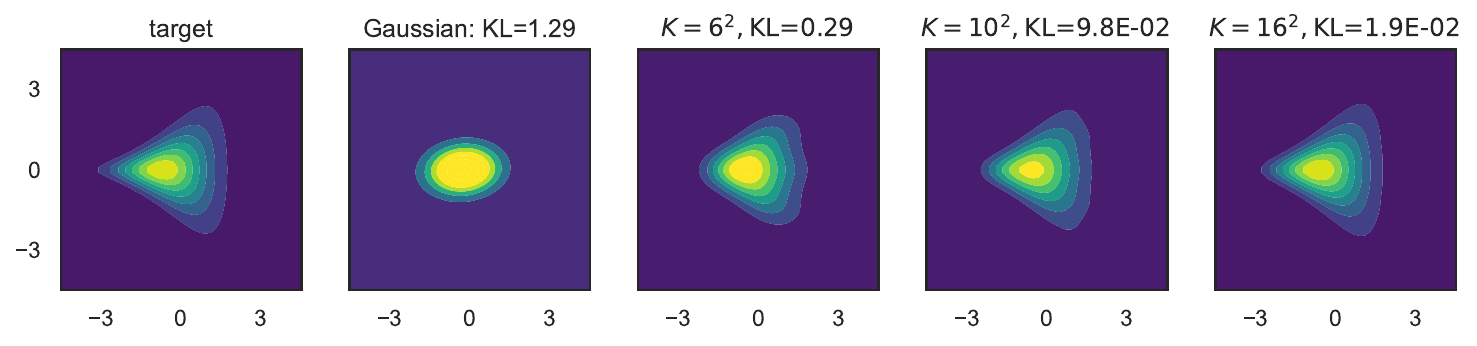}
    \includegraphics[scale=0.44]{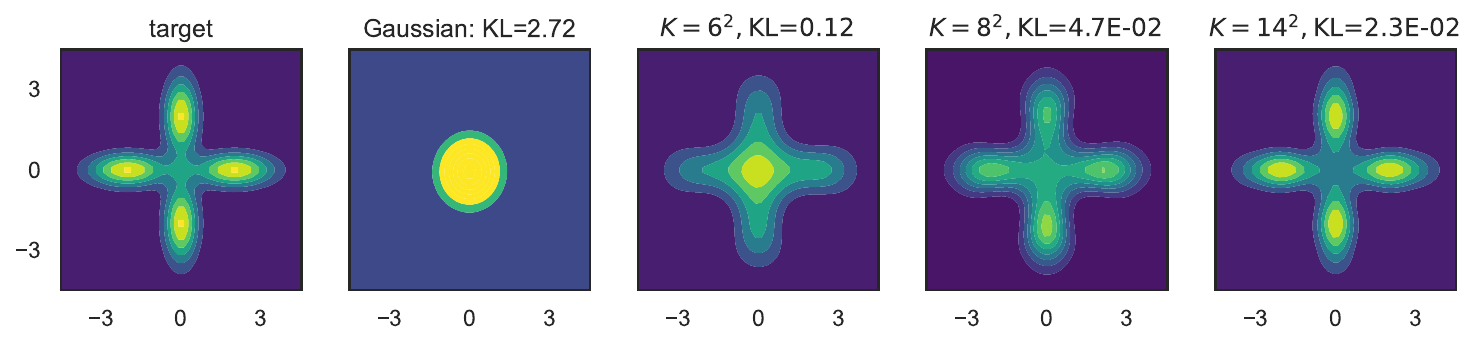}
\caption{2D target functions (column 1): a 3-component Gaussian mixture distribution (row 1), a
funnel distribution (row 2), and a cross distribution (row 3).
    We report the $\KL(p;q)$ for the resulting optimal variational distributions obtained
    using score-based VI with a Gaussian variational family (column 2)
    and the EigenVI variational family (columns 3--5),
    where $K\!=\!K_1K_2$.
    }
\vspace{-10pt}
\label{fig:2dtargets}
\end{figure*}

We first demonstrate how higher-order expansions of the variational family
yield more accurate approximations on a range of 2D non-Gaussian target distributions
(\Cref{fig:2dtargets});
see \Cref{ssec-2dsynthetic} for details.
We report an estimate of $\KL(p;q)$ above each variational approximation.
The Gaussian variational approximation is fit using batch and match VI~\citep{cai2024},
which minimizes a score-based divergence.
For EigenVI,
the target distributions were not standardized before fitting EigenVI
(we compare the costs of the methods in \Cref{fig:2dtargets-metrics}),
and the total number of basis functions is $K\!=\!K_1 K_2$.

\subsection{Non-Gaussianity:\ varying skew and tails in the sinh-arcsinh distribution}

\begin{figure*}[t]
    \centering
    \begin{subfigure}[b]{\linewidth}
    \centering
    \includegraphics[scale=0.32]{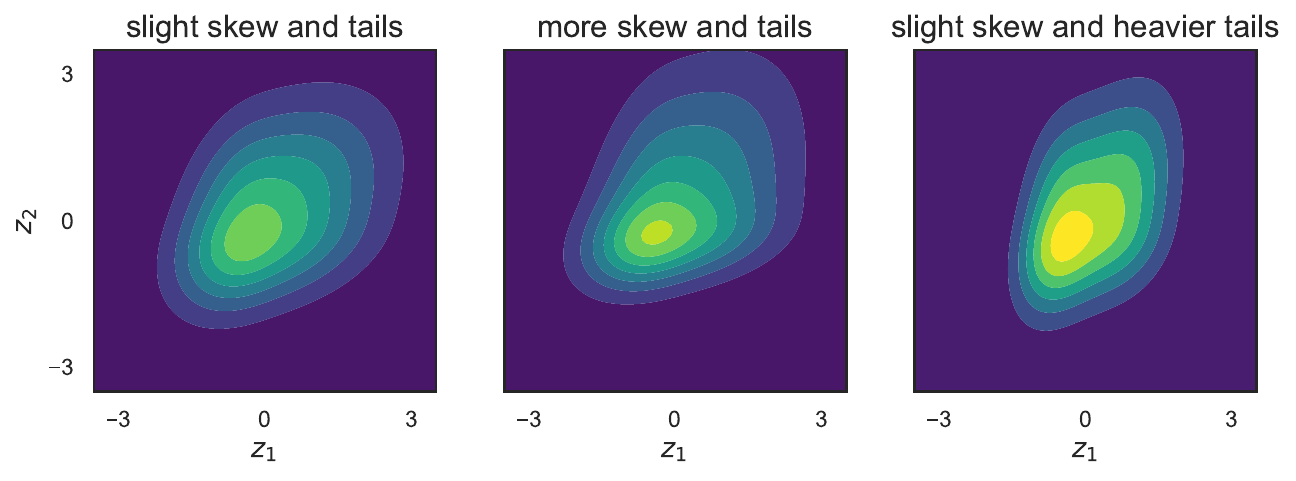}
    \includegraphics[scale=0.32]{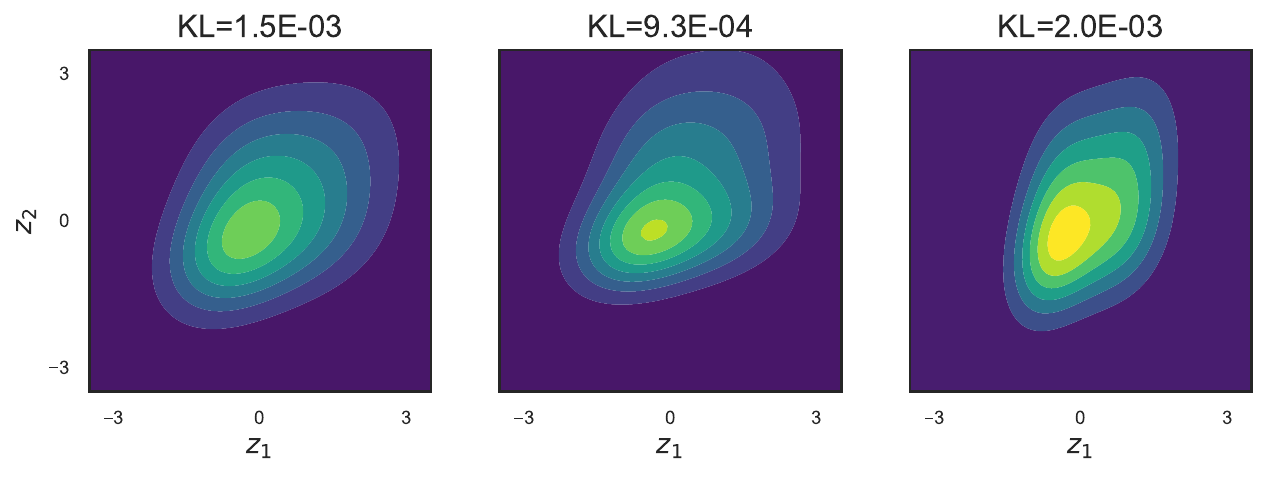}
    \caption{Example 2D targets (left)  varying the skew $s$ or tail weight $\tau$
    components and their EigenVI fits (right).
    }
    \label{fig:sinh2dtargets}
\hspace{10pt}
    \end{subfigure}
    \begin{subfigure}[b]{\linewidth}
        \centering
    \includegraphics[scale=0.21]{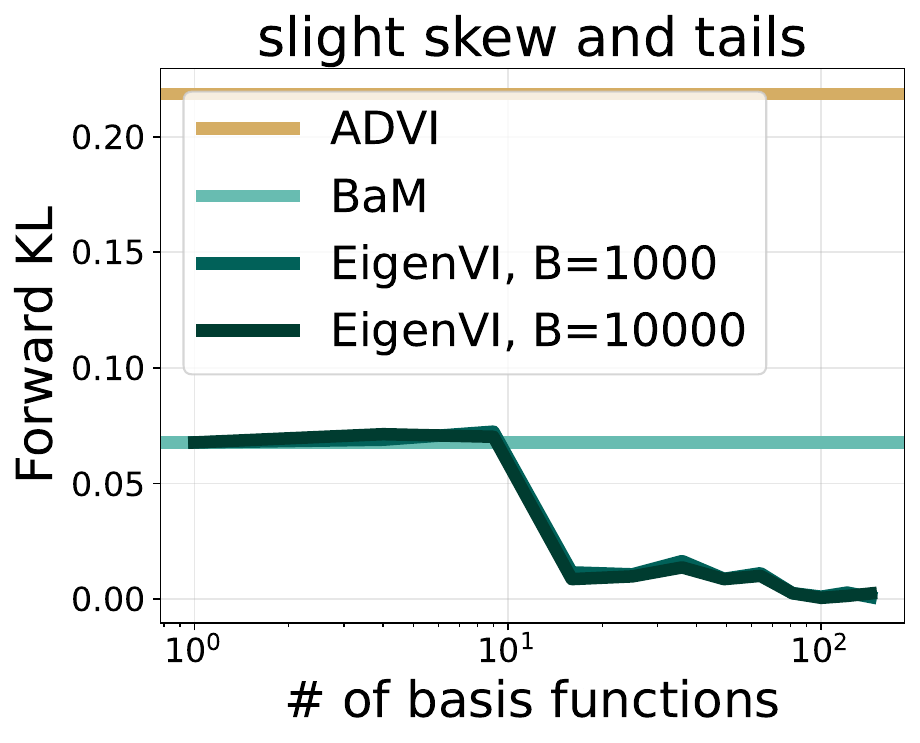}
    \includegraphics[scale=0.21]{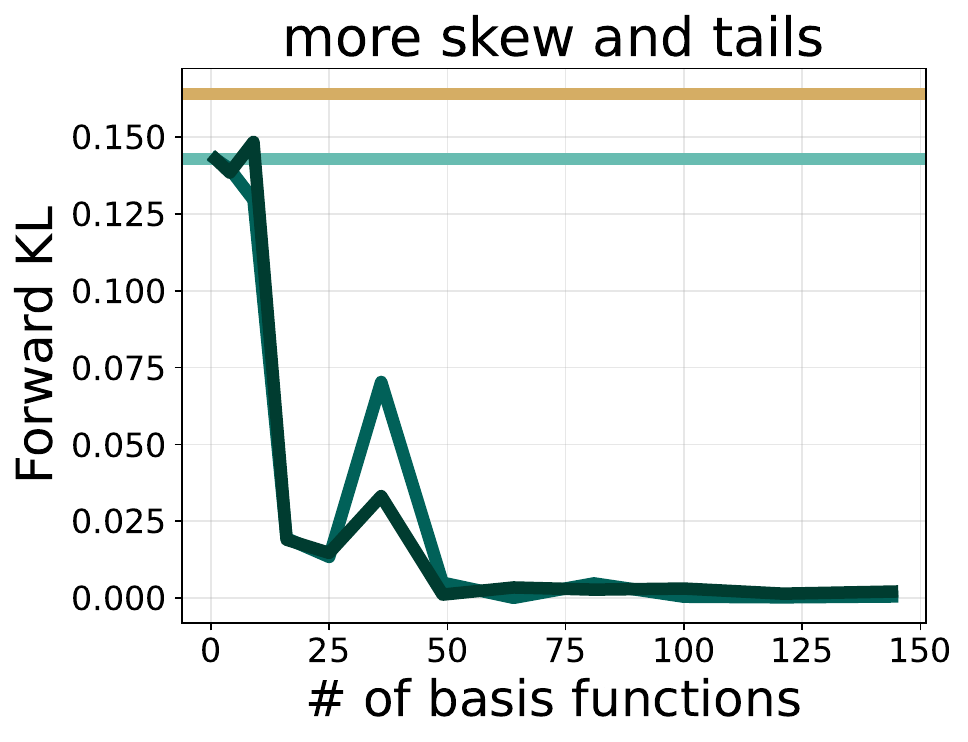}
    \includegraphics[scale=0.21]{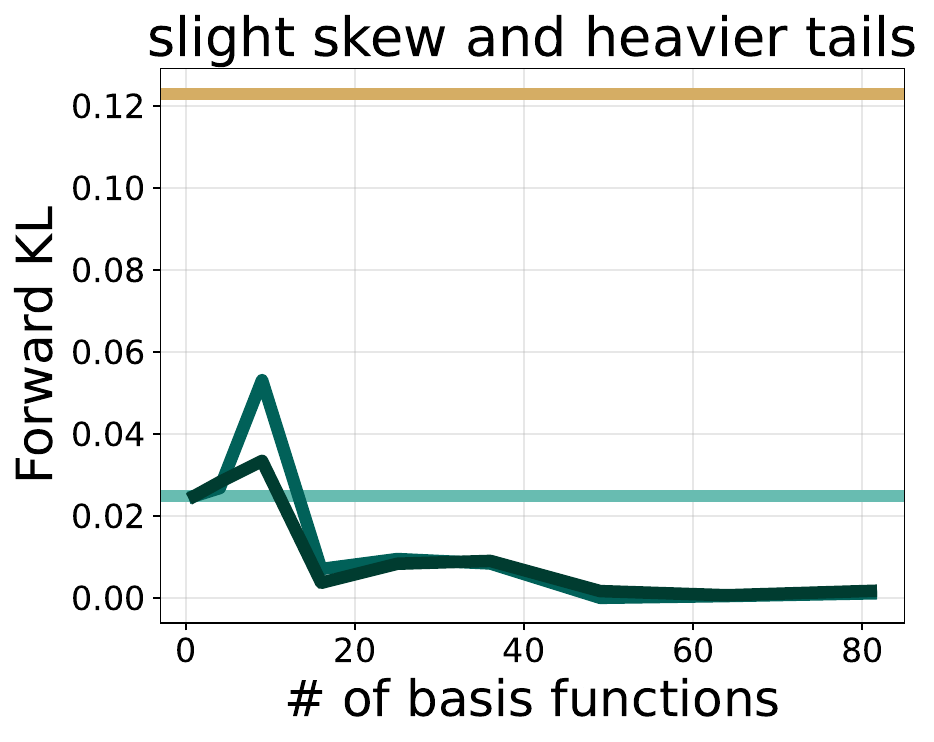}
\caption{$D=2$}
\label{subfig:sinh2D}
    \end{subfigure}
    \begin{subfigure}[b]{\linewidth}
        \centering
    \includegraphics[scale=0.21]{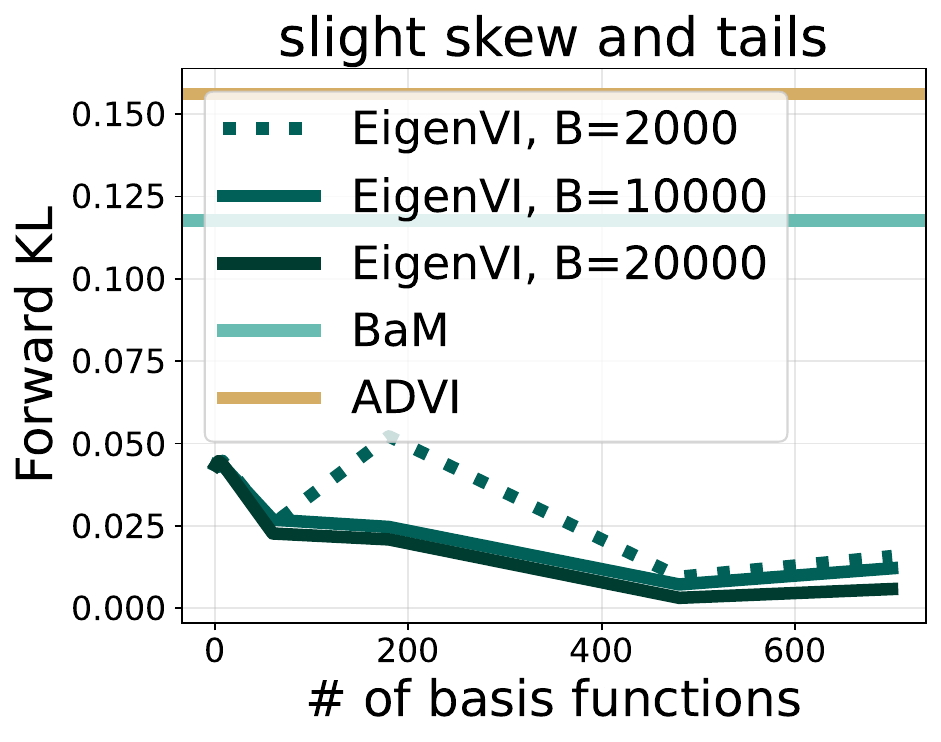}
    \includegraphics[scale=0.21]{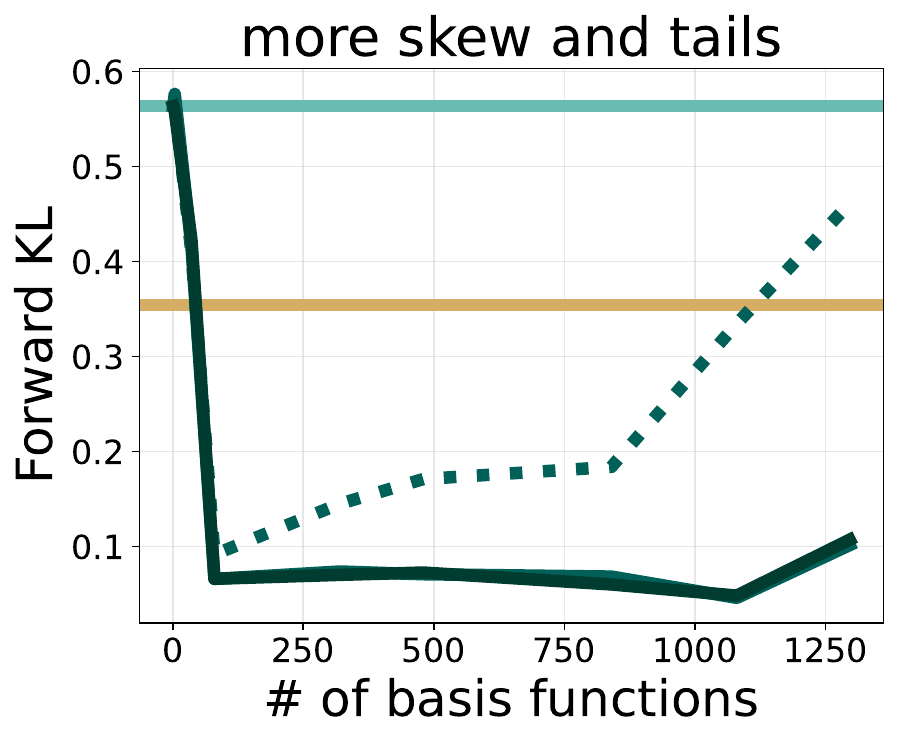}
    \includegraphics[scale=0.21]{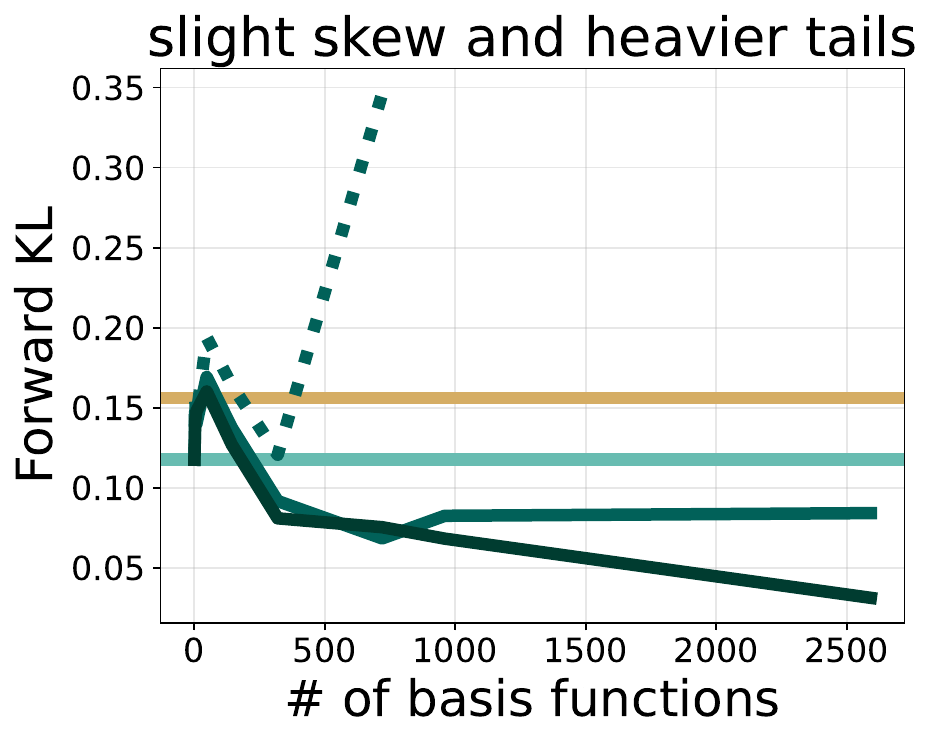}
    \caption{$D=5$}
\label{subfig:sinh5D}
    \end{subfigure}
\caption{Sinh-arcsinh normal distribution synthetic target.
    Panel (a) shows the three targets we consider in 2D, and their resulting EigenVI fit.
    Panel (b) shows measures $\KL(p;q)$ for $D=2$, and panel (c) shows $\KL(p;q)$ for $D=5$;
    the $x$-axis shows the number of basis functions, $K\!=\!\prod_d K_d$.
    }
\vspace{-12pt}
\label{fig:sinhexample}
\end{figure*}

We now consider the  sinh-arcsinh normal distribution \citep{jones2009sinh,jones2019sinh},
which is induced by transforming a multivariate Gaussian using parameters that
control the amount of skew and the weight of the tails.
We construct several targets ($D=2,5$) of
increasing amounts of non-Gaussianity in the skew or the tails of the distribution,
and we refer to these targets as
\emph{slight skew and tails}, \emph{more skew and tails}, and \emph{slight skew and heavier
tails}; see  \Cref{ssec-sinh} for details.
In \Cref{fig:sinh2dtargets},
we visualize the 2D targets and the EigenVI fits along with their forward KLs.
Before applying EigenVI, we standardize the target using a
mean and covariance estimated from batch and match VI \citep{cai2024}.
In \Cref{subfig:sinh2D}, we measure the EigenVI forward KL under varying numbers of samples $B$
and across increasing numbers of basis functions, given by $K\!=\!\prod_{d=1}^D K_d$.
We also present the forward KL resulting from  batch and match VI (BaM) and automatic differentiation VI (ADVI),
which both use Gaussian variational families and are run using the same budget in terms of number gradient
evaluations.
Next we consider similar targets with $D=5$, which are visualized in
in \Cref{fig:5dtargetdensity}, along with the resulting EigenVI variational approximations.
In \Cref{subfig:sinh5D},
we observe greater differences in the number of importance samples
needed to lead to good approximations, especially as the number of basis functions increase.

\subsection{Hierarchical modeling benchmarks from posteriordb}

\begin{figure*}[t]
    \centering
    \begin{subfigure}[b]{\linewidth}
        \centering
        \includegraphics[scale=0.37]{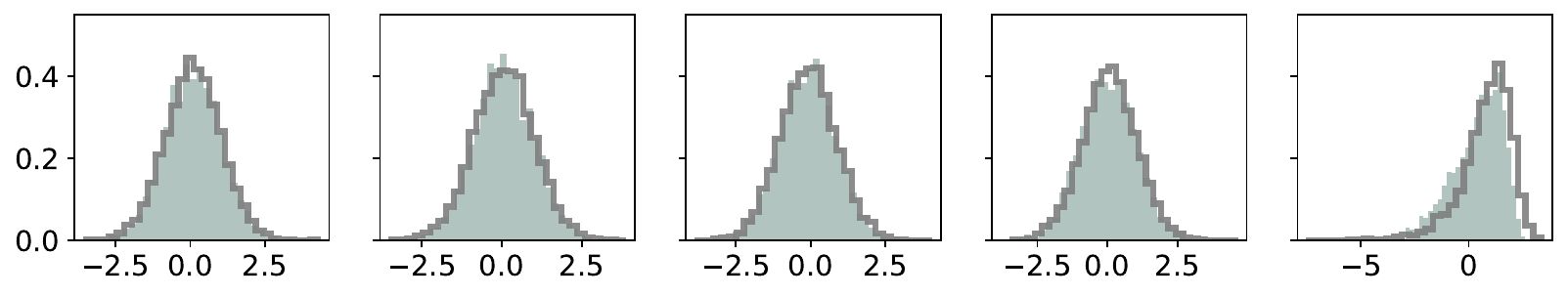}
        \caption{EigenVI with normalized Hermite polynomial family}
    \end{subfigure}
    \begin{subfigure}[b]{\linewidth}
    \centering
        \includegraphics[scale=0.37]{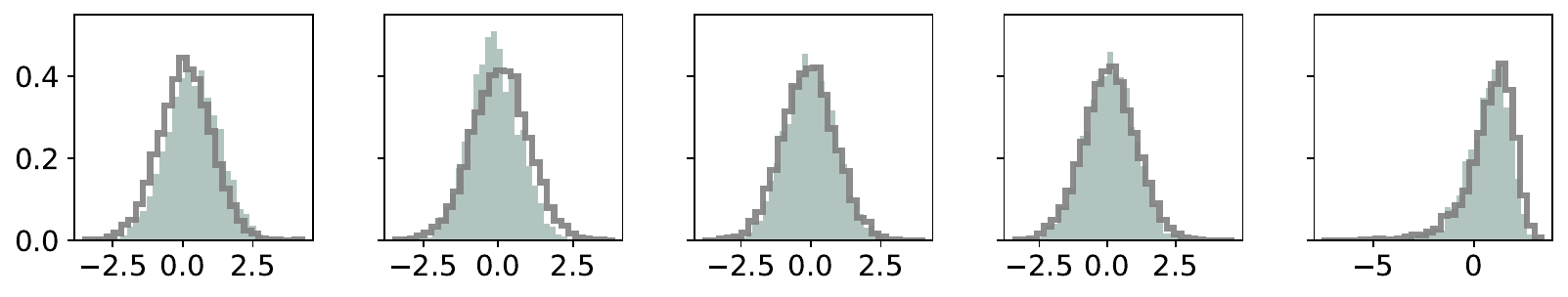}
        \caption{VI with a normalizing flow family}
    \end{subfigure}
    \begin{subfigure}[b]{\linewidth}
    \centering
        \includegraphics[scale=0.37]{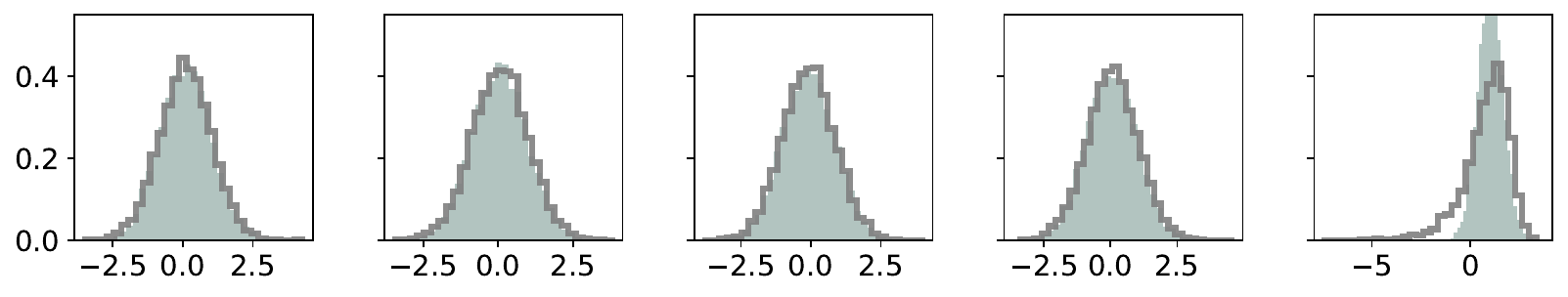}
    \caption{Batch and match VI with a full covariance Gaussian family}
    \vspace{5pt}
    \end{subfigure}
    \begin{subfigure}[b]{0.245\linewidth}
        \centering
        \includegraphics[scale=0.21]{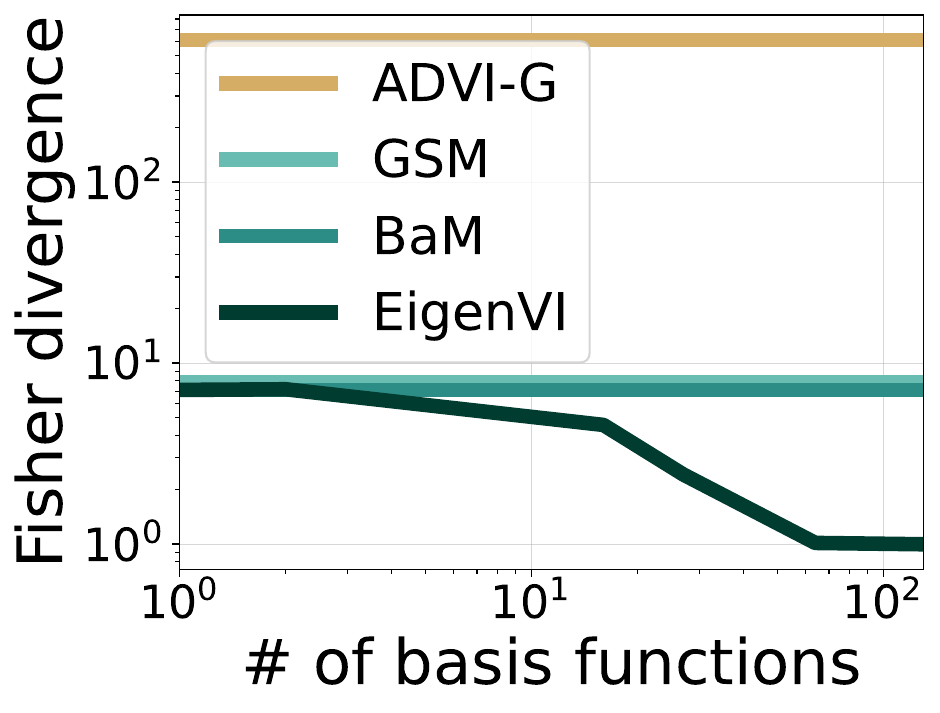}
    \caption{\texttt{kidscore}, $D=3$}
    \end{subfigure}
    \begin{subfigure}[b]{0.245\linewidth}
        \centering
        \includegraphics[scale=0.21]{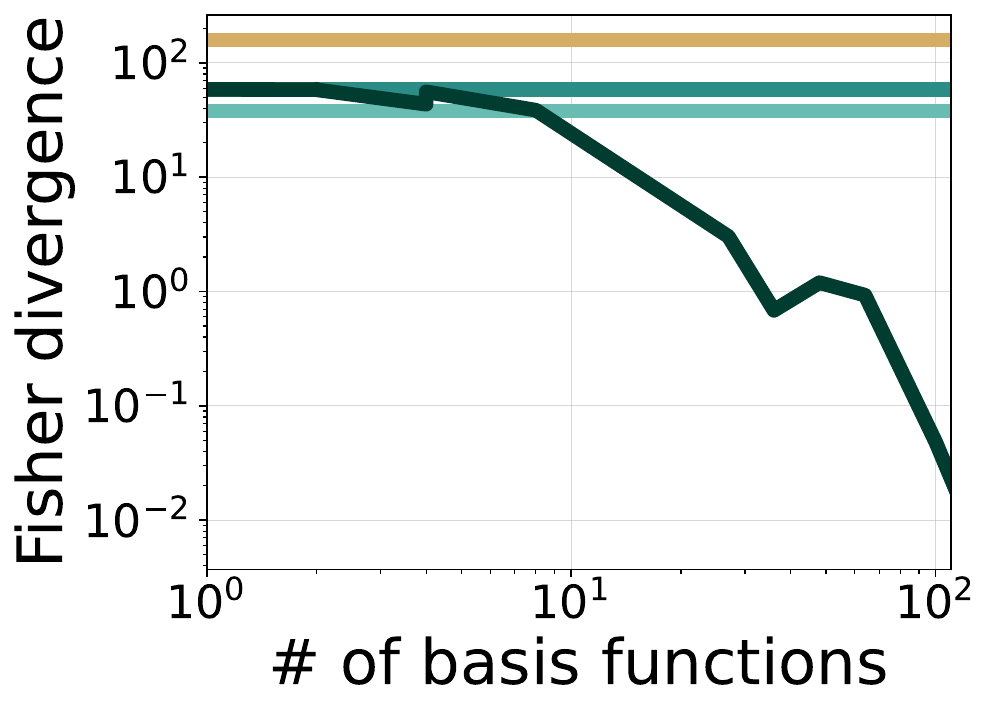}
    \caption{\texttt{sesame}, $D=3$}
    \end{subfigure}
    \begin{subfigure}[b]{0.245\linewidth}
        \centering
        \includegraphics[scale=0.21]{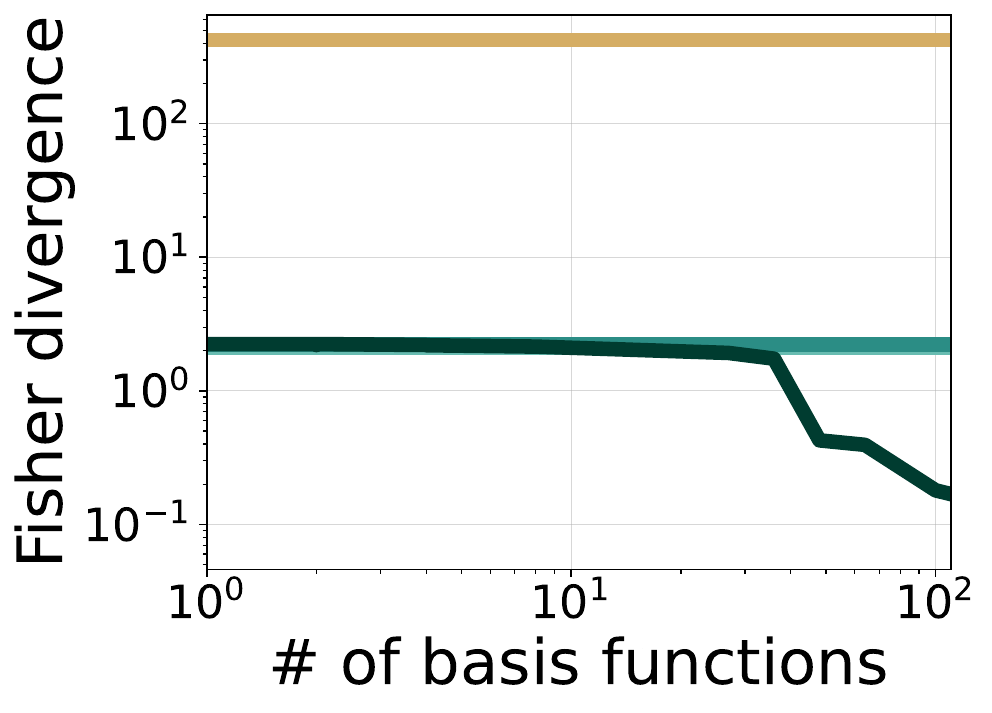}
        \caption{\texttt{gp-regr}, $D=3$
            }
    \end{subfigure}
    \begin{subfigure}[b]{0.245\linewidth}
        \centering
        \includegraphics[scale=0.21]{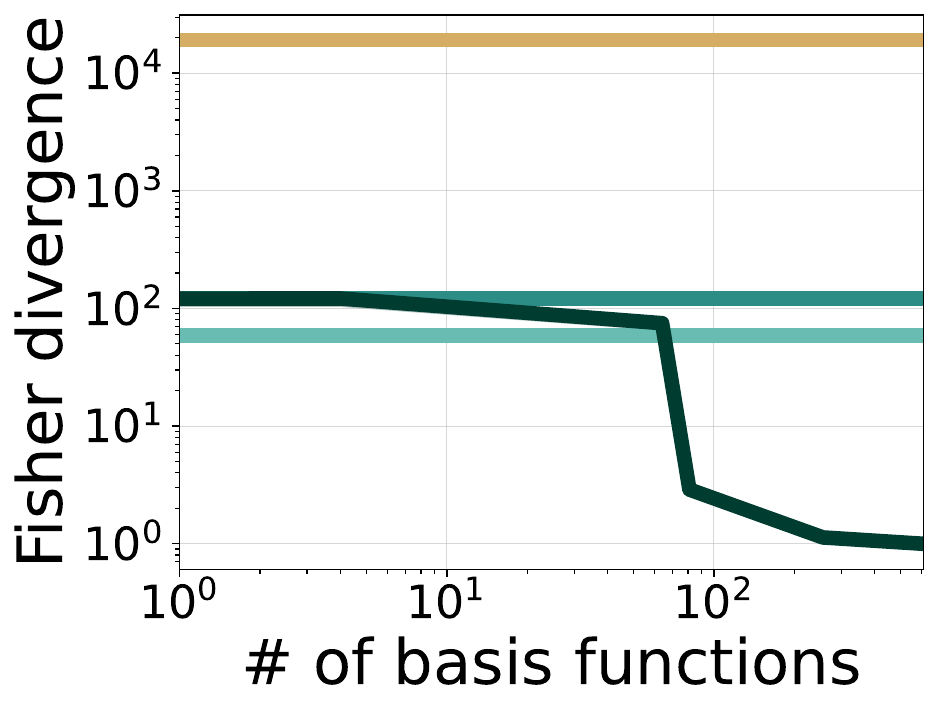}
    \caption{\texttt{logearn}, $D=4$}
    \end{subfigure}
    \begin{subfigure}[b]{0.245\linewidth}
        \centering
        \includegraphics[scale=0.21]{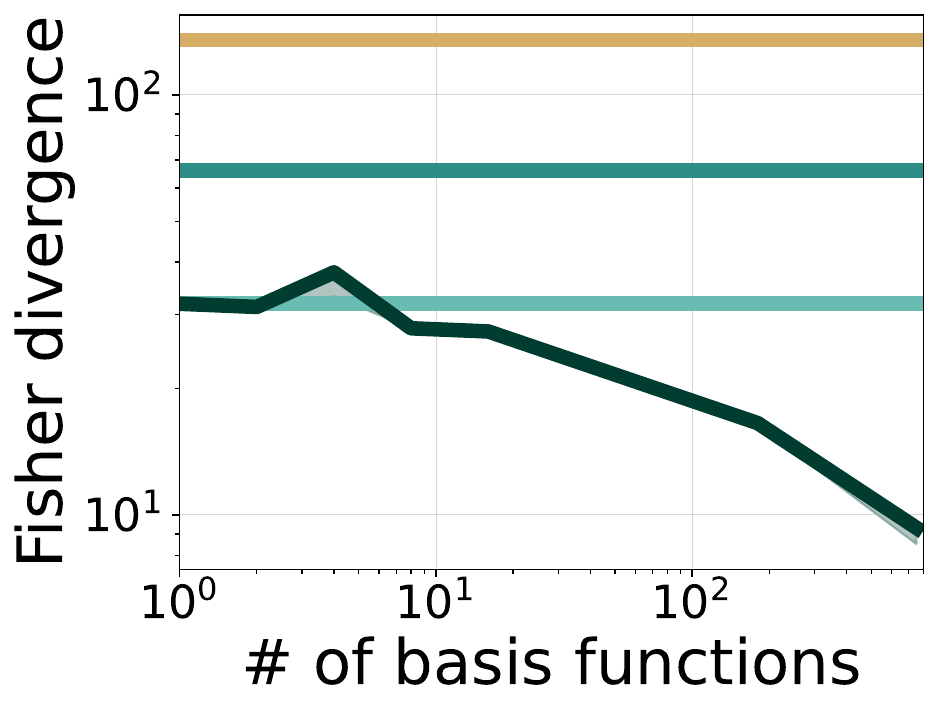}
        \caption{\texttt{garch11}, $D=4$}
    \end{subfigure}
    \begin{subfigure}[b]{0.245\linewidth}
        \centering
        \includegraphics[scale=0.21]{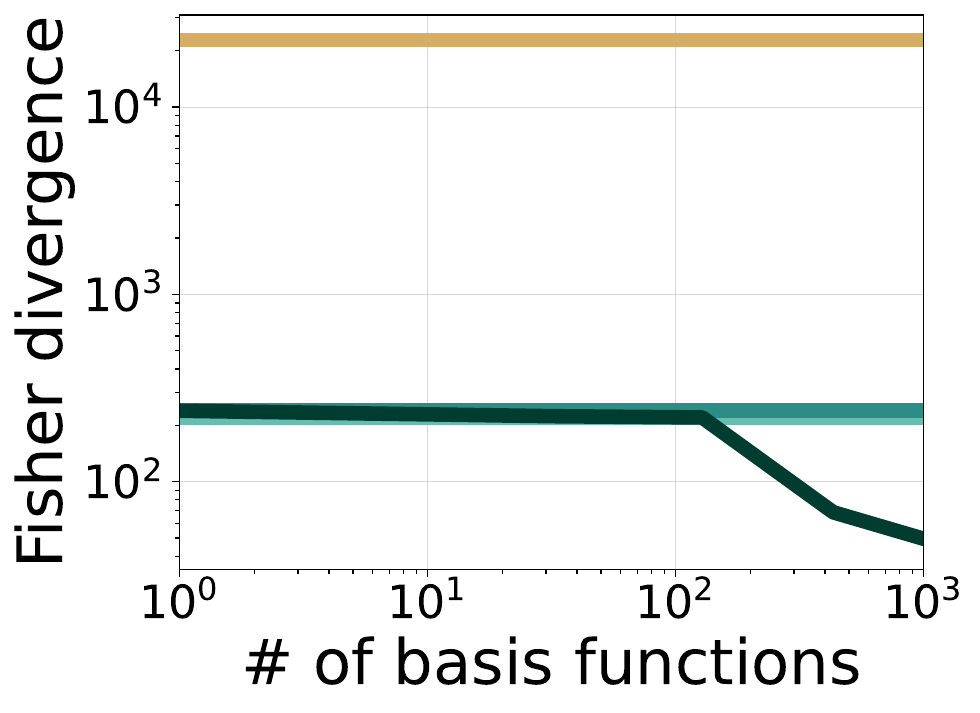}
\caption{\texttt{arK-arK}, $D=7$
            }
    \end{subfigure}
    \begin{subfigure}[b]{0.245\linewidth}
        \includegraphics[scale=0.21]{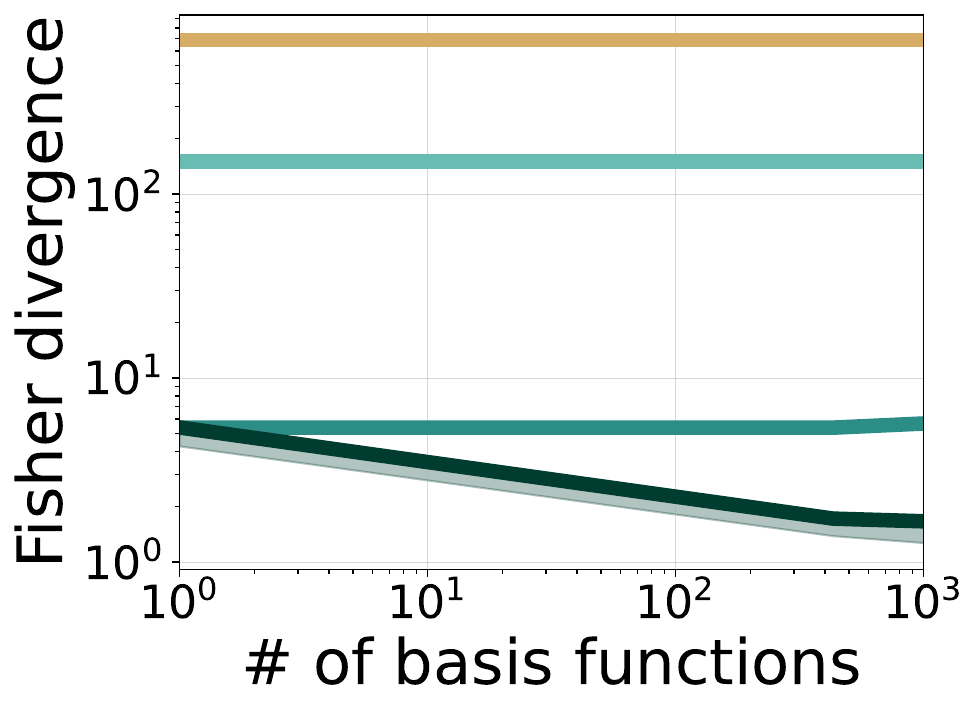}
    \caption{\texttt{logmesquite}, $D=7$ 
            }
    \end{subfigure}
    \begin{subfigure}[b]{0.245\linewidth}
        \includegraphics[scale=0.205]{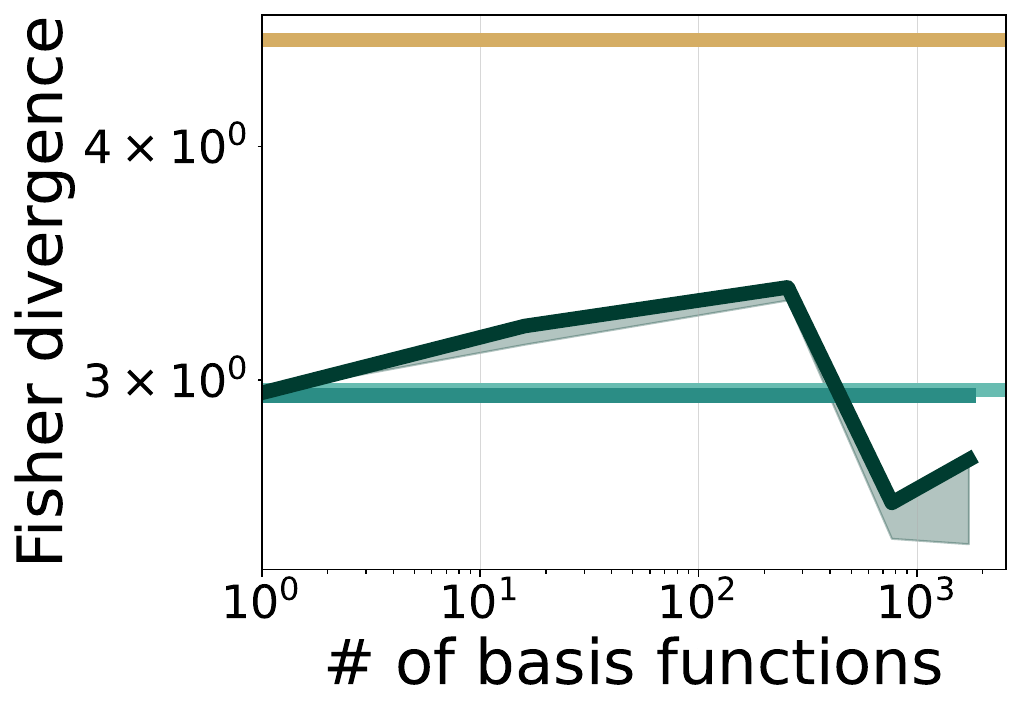}
    \caption{\texttt{8-schools}, $D=10$
            }
    \end{subfigure}
\caption{Results on \texttt{posteriordb} models. Top three rows:
marginal distributions of the even dimensions from \texttt{8-schools}.
Reference samples from HMC are outlined in gray, and
the VI samples are in green.
Bottom two rows: evaluation of methods with the (forward) Fisher divergence.
The $x$-axis shows the number of basis functions, $K\!=\!\prod_{d} K_d$.
Shaded regions represent standard errors computed with respect to 5 random seeds.
}
\label{fig:posteriordb}
\vspace{-10pt}
\end{figure*}

We now evaluate EigenVI on a set of hierarchical Bayesian models
\cite{carpenter2017stan,magnusson2022posteriordb,roualdes2023bridgestan},
which are summarized in \Cref{table:posteriordb}.
The goal is posterior inference: given data  observations $x_{1:N}$,
the posterior of $z$ is
\begin{align}
p(z \given x_{1:N}) \propto p(z)  p(x_{1:N}\given z) =: \rho(z),
\end{align}
where $p(z)$ is the prior and $p(x_{1:N}\given z)$ denotes the likelihood.

We compare EigenVI to
1) automatic differentation VI (ADVI) \citep{kucukelbir2017automatic}, which maximizes the ELBO over a full-covariance Gaussian family
({ADVI}),
2) Gaussian score matching ({GSM}) \citep{modi2023},
a score-based BBVI approach with a full-covariance Gaussian
    family,
and 3)
batch and match VI ({BaM}) \citep{cai2024}, which minimizes a
regularized score-based divergence over a full-covariance Gaussian
family.
In these examples, we standardize the target using either GSM or BaM
before applying EigenVI.

In these models, we do not have access to the target distribution, $p(z \given x_{1:N})$,
only the unnormalized target $\rho$.
Thus, we cannot evaluate an estimate of the forward KL.
Instead, to evaluate the fidelity of the fitted variational distributions,
we compute the empirical Fisher divergence using reference samples from the posterior
obtained via Hamiltonian Monte Carlo (HMC):
\begin{align}
\frac{1}{S} \sum_{s=1}^S \norm{\nabla \log \rho(z^s) - \nabla \log q(z^s)}^2,
    \quad z^s \sim p(z \given x_{1:N}).
\end{align}
Note that this measure is not the objective that EigenVI minimizes; it is analogous to the forward KL
divergence, as the expectation is taken with respect to $p$.
We report the results in \Cref{fig:posteriordb},
computing the Fisher divergence for EigenVI with increasing numbers
of basis functions.
We typically found that with more basis functions,
the scores becomes closer to that of the target.

Finally, we provide a qualitative comparison with real-NVP
normalizing flows (NFs)~\cite{dinh2016density}, a flexible variational
family that is fit by   minimizing the reverse KL.
We found that after tuning the batch-size and learning rate,
NFs generally had a suitable fit.
We visualize the posterior marginals  for a subset of dimensions from \texttt{8schools} in
the top three rows, comparing EigenVI, the NF, and BaM.
Here, we observe that the Gaussian struggles to fit the tails
of this target distribution.
On the other hand, EigenVI provides a competitive fit to the normalizing flow.
In \Cref{ssec:pdb},
we show the full corner plot in \Cref{fig:8schools:corner} and marginals of
the \texttt{garch11} model  in  \Cref{fig:garch11:corner}.




\section{Discussion of limitations and future work}
\label{sec:conclusion}

In this work, we introduced EigenVI, a new approach for score-based variational inference based on orthgonal function expansions.
%
The score-based objective for EigenVI is minimized by solving an eigenvalue problem, and thus this framework provides an alternative to gradient-based methods for BBVI.
Importantly, many computations in EigenVI can be parallelized with respect to the batch of samples,
unlike in iterative methods. We applied EigenVI to many synthetic and real-world targets, and these experiments show that EigenVI provides
a principled way of improving upon Gaussian variational families.

Many future directions remain.
First, the approach described in this paper relies on importance sampling, and thus it may benefit from
more sophisticated methods for adaptive importance sampling. 
Second, it may be useful to construct  variational families from different orthogonal function expansions. Our empirical study focused on the family built from
normalized Hermite polynomials. But this family may require a very high-order expansion to model highly non-Gaussian targets, and such an expansion will be very expensive in high dimensions.
Though this family was sufficient for many of the targets we simulated, others will be crucial for modeling highly non-Gaussian targets.
%
Another direction is to develop variational families whose orthogonal function
expansions scale more favorably with the dimension, perhaps by incorporating low rank structure in the target's covariance.
%
Finally, it would be interesting to explore iterative versions of EigenVI in which each iteration solves a minimum eigenvalue problem on some subsample of data points. With such an approach, EigenVI could potentially be applied to very large-scale problems in Bayesian inference.



\begin{ack}
We thank Bob Carpenter and Yuling Yao for helpful discussions
and  anonymous reviewers for their time and feedback on the paper.
The Flatiron Institute is a division
of the Simons Foundation.
This work was supported in
part by NSF IIS-2127869, NSF DMS-2311108, NSF/DoD PHY-2229929, ONR N00014-17-1-2131,
ONR N00014-15-1-2209, the Simons Foundation, and Open Philanthropy.
\end{ack}

\bibliographystyle{plainnat-mod}
\bibliography{main}

\begin{thebibliography}{53}
\providecommand{\natexlab}[1]{#1}
\providecommand{\url}[1]{\texttt{#1}}
\expandafter\ifx\csname urlstyle\endcsname\relax
  \providecommand{\doi}[1]{doi: #1}\else
  \providecommand{\doi}{doi: \begingroup \urlstyle{rm}\Url}\fi

\bibitem[Abril-Pla et~al.(2023)Abril-Pla, Andreani, Carroll, Dong, Fonnesbeck,
  Kochurov, Kumar, Lao, Luhmann, Martin, et~al.]{abril2023pymc}
O.~Abril-Pla, V.~Andreani, C.~Carroll, L.~Dong, C.~J. Fonnesbeck, M.~Kochurov,
  R.~Kumar, J.~Lao, C.~C. Luhmann, O.~A. Martin, et~al.
\newblock {PyMC}: a modern, and comprehensive probabilistic programming
  framework in {P}ython.
\newblock \emph{PeerJ Computer Science}, 9:\penalty0 e1516, 2023.

\bibitem[Agrawal et~al.(2020)Agrawal, Sheldon, and Domke]{agrawal2020advances}
A.~Agrawal, D.~R. Sheldon, and J.~Domke.
\newblock Advances in black-box {VI}: Normalizing flows, importance weighting,
  and optimization.
\newblock \emph{Advances in Neural Information Processing Systems}, 33, 2020.

\bibitem[Berg et~al.(2018)Berg, Hasenclever, Tomczak, and
  Welling]{berg2018sylvester}
R.~v.~d. Berg, L.~Hasenclever, J.~M. Tomczak, and M.~Welling.
\newblock Sylvester normalizing flows for variational inference.
\newblock \emph{Uncertainty in Artificial Intelligence}, 2018.

\bibitem[Bingham et~al.(2019)Bingham, Chen, Jankowiak, Obermeyer, Pradhan,
  Karaletsos, Singh, Szerlip, Horsfall, and Goodman]{bingham2019pyro}
E.~Bingham, J.~P. Chen, M.~Jankowiak, F.~Obermeyer, N.~Pradhan, T.~Karaletsos,
  R.~Singh, P.~Szerlip, P.~Horsfall, and N.~D. Goodman.
\newblock Pyro: Deep universal probabilistic programming.
\newblock \emph{The Journal of Machine Learning Research}, 20\penalty0
  (1):\penalty0 973--978, 2019.

\bibitem[Blei et~al.(2017)Blei, Kucukelbir, and McAuliffe]{blei2017vi}
D.~M. Blei, A.~Kucukelbir, and J.~D. McAuliffe.
\newblock Variational inference: A review for statisticians.
\newblock \emph{Journal of the American Statistical Association}, 112\penalty0
  (518):\penalty0 859--877, 2017.

\bibitem[Cai et~al.(2024)Cai, Modi, Pillaud-Vivien, Margossian, Gower, Blei,
  and Saul]{cai2024}
D.~Cai, C.~Modi, L.~Pillaud-Vivien, C.~Margossian, R.~Gower, D.~Blei, and
  L.~Saul.
\newblock Batch and match: black-box variational inference with a score-based
  divergence.
\newblock In \emph{International Conference on Machine Learning}, 2024.

\bibitem[Carpenter et~al.(2017)Carpenter, Gelman, Hoffman, Lee, Goodrich,
  Betancourt, Brubaker, Guo, Li, and Riddell]{carpenter2017stan}
B.~Carpenter, A.~Gelman, M.~D. Hoffman, D.~Lee, B.~Goodrich, M.~Betancourt,
  M.~Brubaker, J.~Guo, P.~Li, and A.~Riddell.
\newblock Stan: A probabilistic programming language.
\newblock \emph{Journal of Statistical Software}, 76\penalty0 (1):\penalty0
  1--32, 2017.

\bibitem[Courant and Hilbert(1924)]{courant1924methoden}
R.~Courant and D.~Hilbert.
\newblock \emph{Methoden der Mathematischen Physik}, volume~1.
\newblock Julius Springer, Berlin, 1924.

\bibitem[Dai et~al.(2019)Dai, Dai, Gretton, Song, Schuurmans, and
  He]{dai2019kernel}
B.~Dai, H.~Dai, A.~Gretton, L.~Song, D.~Schuurmans, and N.~He.
\newblock Kernel exponential family estimation via doubly dual embedding.
\newblock In \emph{International Conference on Artificial Intelligence and
  Statistics}. PMLR, 2019.

\bibitem[Dhaka et~al.(2020)Dhaka, Catalina, Andersen, Magnusson, Huggins, and
  Vehtari]{dhaka2020robust}
A.~K. Dhaka, A.~Catalina, M.~R. Andersen, M.~Magnusson, J.~Huggins, and
  A.~Vehtari.
\newblock Robust, accurate stochastic optimization for variational inference.
\newblock \emph{Advances in Neural Information Processing Systems}, 33, 2020.

\bibitem[Dhaka et~al.(2021)Dhaka, Catalina, Welandawe, Andersen, Huggins, and
  Vehtari]{dhaka2021challenges}
A.~K. Dhaka, A.~Catalina, M.~Welandawe, M.~R. Andersen, J.~Huggins, and
  A.~Vehtari.
\newblock Challenges and opportunities in high dimensional variational
  inference.
\newblock \emph{Advances in Neural Information Processing Systems}, 34, 2021.

\bibitem[Dinh et~al.(2017)Dinh, Sohl-Dickstein, and Bengio]{dinh2016density}
L.~Dinh, J.~Sohl-Dickstein, and S.~Bengio.
\newblock Density estimation using real {NVP}.
\newblock \emph{International Conference on Learning Representations}, 2017.

\bibitem[Ge et~al.(2018)Ge, Xu, and Ghahramani]{ge2018turing}
H.~Ge, K.~Xu, and Z.~Ghahramani.
\newblock Turing: a language for flexible probabilistic inference.
\newblock In \emph{International Conference on Artificial Intelligence and
  Statistics}. PMLR, 2018.

\bibitem[Gershman et~al.(2012)Gershman, Hoffman, and
  Blei]{gershman2012nonparametric}
S.~Gershman, M.~Hoffman, and D.~Blei.
\newblock Nonparametric variational inference.
\newblock \emph{International Conference on Machine Learning}, 2012.

\bibitem[Giordano et~al.(2024)Giordano, Ingram, and
  Broderick]{giordano2023black}
R.~Giordano, M.~Ingram, and T.~Broderick.
\newblock Black box variational inference with a deterministic objective:
  Faster, more accurate, and even more black box.
\newblock \emph{Journal of Machine Learning Research}, 25\penalty0
  (18):\penalty0 1--39, 2024.

\bibitem[Griffiths and Schroeter(2018)]{griffiths2018introduction}
D.~J. Griffiths and D.~F. Schroeter.
\newblock \emph{Introduction to Quantum Mechanics}.
\newblock Cambridge University Press, 2018.

\bibitem[Guo et~al.(2016)Guo, Wang, Fan, Broderick, and
  Dunson]{guo2016boosting}
F.~Guo, X.~Wang, K.~Fan, T.~Broderick, and D.~B. Dunson.
\newblock Boosting variational inference.
\newblock \emph{arXiv preprint arXiv:1611.05559}, 2016.

\bibitem[Hyv{\"a}rinen(2005)]{hyvarinen2005estimation}
A.~Hyv{\"a}rinen.
\newblock Estimation of non-normalized statistical models by score matching.
\newblock \emph{Journal of Machine Learning Research}, 6\penalty0 (4), 2005.

\bibitem[Jones and Pewsey(2009)]{jones2009sinh}
C.~Jones and A.~Pewsey.
\newblock Sinh-arcsinh distributions.
\newblock \emph{Biometrika}, 96\penalty0 (4):\penalty0 761--780, 2009.

\bibitem[Jones and Pewsey(2019)]{jones2019sinh}
C.~Jones and A.~Pewsey.
\newblock The sinh-arcsinh normal distribution.
\newblock \emph{Significance}, 16\penalty0 (2):\penalty0 6--7, 2019.

\bibitem[Jordan et~al.(1999)Jordan, Ghahramani, Jaakkola, and
  Saul]{jordan1999vi}
M.~I. Jordan, Z.~Ghahramani, T.~S. Jaakkola, and L.~K. Saul.
\newblock An introduction to variational methods for graphical models.
\newblock \emph{Machine Learning}, 37:\penalty0 183--233, 1999.

\bibitem[Kim and Bengio(2016)]{kim2016deep}
T.~Kim and Y.~Bengio.
\newblock Deep directed generative models with energy-based probability
  estimation.
\newblock \emph{arXiv preprint arXiv:1606.03439}, 2016.

\bibitem[Kingma and Welling(2014)]{kingma2013auto}
D.~P. Kingma and M.~Welling.
\newblock Auto-encoding variational {B}ayes.
\newblock In \emph{International Conference on Learning Representations}, 2014.

\bibitem[Kingma et~al.(2016)Kingma, Salimans, Jozefowicz, Chen, Sutskever, and
  Welling]{kingma2016improved}
D.~P. Kingma, T.~Salimans, R.~Jozefowicz, X.~Chen, I.~Sutskever, and
  M.~Welling.
\newblock Improved variational inference with inverse autoregressive flow.
\newblock \emph{Advances in Neural Information Processing Systems}, 29, 2016.

\bibitem[Kobyzev et~al.(2020)Kobyzev, Prince, and
  Brubaker]{kobyzev2020normalizing}
I.~Kobyzev, S.~J. Prince, and M.~A. Brubaker.
\newblock Normalizing flows: An introduction and review of current methods.
\newblock \emph{IEEE Transactions on Pattern Analysis and Machine
  Intelligence}, 43\penalty0 (11):\penalty0 3964--3979, 2020.

\bibitem[K{\"o}hler et~al.(2021)K{\"o}hler, Kr{\"a}mer, and
  No{\'e}]{kohler2021smooth}
J.~K{\"o}hler, A.~Kr{\"a}mer, and F.~No{\'e}.
\newblock Smooth normalizing flows.
\newblock \emph{Advances in Neural Information Processing Systems}, 34, 2021.

\bibitem[Kucukelbir et~al.(2017)Kucukelbir, Tran, Ranganath, Gelman, and
  Blei]{kucukelbir2017automatic}
A.~Kucukelbir, D.~Tran, R.~Ranganath, A.~Gelman, and D.~M. Blei.
\newblock Automatic differentiation variational inference.
\newblock \emph{Journal of Machine Learning Research}, 2017.

\bibitem[Lawson et~al.(2019)Lawson, Tucker, Dai, and
  Ranganath]{lawson2019energy}
J.~Lawson, G.~Tucker, B.~Dai, and R.~Ranganath.
\newblock Energy-inspired models: Learning with sampler-induced distributions.
\newblock \emph{Advances in Neural Information Processing Systems}, 32, 2019.

\bibitem[LeCun et~al.(2006)LeCun, Chopra, Hadsell, Ranzato, and
  Huang]{lecun2006tutorial}
Y.~LeCun, S.~Chopra, R.~Hadsell, M.~Ranzato, and F.~Huang.
\newblock A tutorial on energy-based learning.
\newblock \emph{Predicting Structured Data}, 1\penalty0 (0), 2006.

\bibitem[Lehoucq et~al.(1998)Lehoucq, Sorensen, and Yang]{arpack1998}
R.~B. Lehoucq, D.~C. Sorensen, and C.~Yang.
\newblock \emph{{ARPACK Users' Guide: Solution of Large-Scale Eigenvalue
  Problems with Implicitly Restarted Arnoldi Methods}}.
\newblock SIAM, 1998.
\newblock Available at \url{http://www.caam.rice.edu/software/ARPACK/}.

\bibitem[Liu and Wang(2016)]{liu2016stein}
Q.~Liu and D.~Wang.
\newblock Stein variational gradient descent:\ a general purpose {B}ayesian
  inference algorithm.
\newblock \emph{Advances in Neural Information Processing Systems}, 29, 2016.

\bibitem[Locatello et~al.(2018)Locatello, Dresdner, Khanna, Valera, and
  R{\"a}tsch]{locatello2018boosting}
F.~Locatello, G.~Dresdner, R.~Khanna, I.~Valera, and G.~R{\"a}tsch.
\newblock Boosting black box variational inference.
\newblock \emph{Advances in Neural Information Processing Systems}, 31, 2018.

\bibitem[Loconte et~al.(2024)Loconte, Sladek, Mengel, Trapp, Solin, Gillis, and
  Vergari]{loconte2024subtractive}
L.~Loconte, A.~M. Sladek, S.~Mengel, M.~Trapp, A.~Solin, N.~Gillis, and
  A.~Vergari.
\newblock Subtractive mixture models via squaring: Representation and learning.
\newblock In \emph{International Conference on Learning Representations}, 2024.

\bibitem[Louizos and Welling(2017)]{louizos2017multiplicative}
C.~Louizos and M.~Welling.
\newblock Multiplicative normalizing flows for variational {B}ayesian neural
  networks.
\newblock In \emph{International Conference on Machine Learning}. PMLR, 2017.

\bibitem[Magnusson et~al.(2022)Magnusson, Bürkner, and
  Vehtari]{magnusson2022posteriordb}
M.~Magnusson, P.~Bürkner, and A.~Vehtari.
\newblock posteriordb: a set of posteriors for {B}ayesian inference and
  probabilistic programming.
\newblock \url{https://github.com/stan-dev/posteriordb}, 2022.

\bibitem[Miller et~al.(2017)Miller, Foti, and Adams]{miller2017variational}
A.~C. Miller, N.~J. Foti, and R.~P. Adams.
\newblock Variational boosting: Iteratively refining posterior approximations.
\newblock In \emph{International Conference on Machine Learning}, pages
  2420--2429. PMLR, 2017.

\bibitem[Modi et~al.(2023)Modi, Margossian, Yao, Gower, Blei, and
  Saul]{modi2023}
C.~Modi, C.~Margossian, Y.~Yao, R.~Gower, D.~Blei, and L.~Saul.
\newblock Variational inference with {G}aussian score matching.
\newblock \emph{Advances in Neural Information Processing Systems}, 36, 2023.

\bibitem[Novikov et~al.(2021)Novikov, Panov, and Oseledets]{novikov2021tensor}
G.~S. Novikov, M.~E. Panov, and I.~V. Oseledets.
\newblock Tensor-train density estimation.
\newblock In \emph{Uncertainty in Artificial Intelligence}, pages 1321--1331.
  PMLR, 2021.

\bibitem[Papamakarios et~al.(2021)Papamakarios, Nalisnick, Rezende, Mohamed,
  and Lakshminarayanan]{papamakarios2021normalizing}
G.~Papamakarios, E.~Nalisnick, D.~J. Rezende, S.~Mohamed, and
  B.~Lakshminarayanan.
\newblock Normalizing flows for probabilistic modeling and inference.
\newblock \emph{Journal of Machine Learning Research}, 22\penalty0
  (57):\penalty0 1--64, 2021.

\bibitem[Ranganath et~al.(2014)Ranganath, Gerrish, and
  Blei]{ranganath2014black}
R.~Ranganath, S.~Gerrish, and D.~Blei.
\newblock Black box variational inference.
\newblock In \emph{Artificial Intelligence and Statistics}, pages 814--822.
  PMLR, 2014.

\bibitem[Rezende and Mohamed(2015)]{rezende2015variational}
D.~Rezende and S.~Mohamed.
\newblock Variational inference with normalizing flows.
\newblock In \emph{International Conference on Machine Learning}. PMLR, 2015.

\bibitem[Roualdes et~al.(2023)Roualdes, Ward, Axen, and
  Carpenter]{roualdes2023bridgestan}
E.~Roualdes, B.~Ward, S.~Axen, and B.~Carpenter.
\newblock Bridge{S}tan: Efficient in-memory access to {S}tan programs through
  {P}ython, {J}ulia, and {R}.
\newblock \url{https://github.com/roualdes/bridgestan}, 2023.

\bibitem[Salvatier et~al.(2016)Salvatier, Wiecki, and
  Fonnesbeck]{salvatier2016probabilistic}
J.~Salvatier, T.~V. Wiecki, and C.~Fonnesbeck.
\newblock Probabilistic programming in {P}ython using {PyMC3}.
\newblock \emph{PeerJ Computer Science}, 2:\penalty0 e55, 2016.

\bibitem[Titsias and L{\'a}zaro-Gredilla(2014)]{titsias2014doubly}
M.~Titsias and M.~L{\'a}zaro-Gredilla.
\newblock Doubly stochastic variational {B}ayes for non-conjugate inference.
\newblock In \emph{International Conference on Machine Learning}. PMLR, 2014.

\bibitem[Wainwright et~al.(2008)Wainwright, Jordan,
  et~al.]{wainwright2008graphical}
M.~J. Wainwright, M.~I. Jordan, et~al.
\newblock Graphical models, exponential families, and variational inference.
\newblock \emph{Foundations and Trends{\textregistered} in Machine Learning},
  1\penalty0 (1--2):\penalty0 1--305, 2008.

\bibitem[Wang et~al.(2024)Wang, Geffner, and Domke]{wang2022dual}
X.~Wang, T.~Geffner, and J.~Domke.
\newblock Dual control variate for faster black-box variational inference.
\newblock In \emph{International Conference on Artificial Intelligence and
  Statistics}, 2024.

\bibitem[Yang et~al.(2019)Yang, Martin, and Bondell]{yang2019variational}
Y.~Yang, R.~Martin, and H.~Bondell.
\newblock Variational approximations using {F}isher divergence.
\newblock \emph{arXiv preprint arXiv:1905.05284}, 2019.

\bibitem[Yu and Zhang(2023)]{yu2023semiimplicit}
L.~Yu and C.~Zhang.
\newblock Semi-implicit variational inference via score matching.
\newblock In \emph{International Conference on Learning Representations}, 2023.

\bibitem[Zeghal et~al.(2022)Zeghal, Lanusse, Boucaud, Remy, and
  Aubourg]{Zeghal2022npe}
J.~Zeghal, F.~Lanusse, A.~Boucaud, B.~Remy, and E.~Aubourg.
\newblock {Neural Posterior Estimation with Differentiable Simulators}.
\newblock In \emph{{International Conference on Machine Learning Conference}},
  2022.

\bibitem[Zhang et~al.(2018)Zhang, Shahbaba, and Zhao]{zhang2018variational}
C.~Zhang, B.~Shahbaba, and H.~Zhao.
\newblock Variational {H}amiltonian {M}onte {C}arlo via score matching.
\newblock \emph{Bayesian Analysis}, 13\penalty0 (2):\penalty0 485, 2018.

\bibitem[Zhang et~al.(2022)Zhang, Carpenter, Gelman, and
  Vehtari]{zhang2022pathfinder}
L.~Zhang, B.~Carpenter, A.~Gelman, and A.~Vehtari.
\newblock Pathfinder: Parallel quasi-newton variational inference.
\newblock \emph{Journal of Machine Learning Research}, 23\penalty0
  (306):\penalty0 1--49, 2022.

\bibitem[Zhu et~al.(1998)Zhu, Wu, and Mumford]{zhu1998filters}
S.~C. Zhu, Y.~Wu, and D.~Mumford.
\newblock Filters, random fields and maximum entropy ({FRAME}): Towards a
  unified theory for texture modeling.
\newblock \emph{International Journal of Computer Vision}, 27:\penalty0
  107--126, 1998.

\bibitem[Zoltowski et~al.(2021)Zoltowski, Cai, and Adams]{zoltowski2021slice}
D.~Zoltowski, D.~Cai, and R.~P. Adams.
\newblock Slice sampling reparameterization gradients.
\newblock \emph{Advances in Neural Information Processing Systems},
  34:\penalty0 23532--23544, 2021.

\end{thebibliography}


\newpage

\appendix

\numberwithin{equation}{section}
\numberwithin{figure}{section}
\numberwithin{table}{section}



\section{Sampling from orthogonal function expansions}
\label{app:sampling}

In this appendix we show how to sample from a density on $\mathbb{R}^D$ constructed from a Cartesian product of orthogonal function expansions. Specifically, we assume that the density is of the form
\begin{equation}
    q(z_1,z_2,\ldots,z_D) = \left(\sum_{k_1=1}^{K_1} \cdots \sum_{k_D=1}^{K_D} \alpha_{k_1 k_2 \ldots k_D}\phi_{k_1}(z_1)\phi_{k_2}(z_2)\cdots\phi_{k_D}(z_D)\right)^2,
\end{equation}
where $\{\phi_{k}(\cdot)\}_{k=1}^\infty$ define a family of orthonormal functions on $\mathbb{R}$ and where the density is normalized by requiring that
\begin{equation}
\sum_{k_1 k_2\ldots k_D} \alpha_{k_1 k_2\ldots k_D}^2=1.
\end{equation}
To draw samples from this density, we describe a sequential procedure based on inverse transform sampling. In particular, we obtain a sample $z\in\R^D$ by the sequence of draws
\begin{align}
\label{eq:draw1}
z_1 & \sim  q(z_1),  \\
z_2 & \sim  q(z_2|z_1), \\
    \vdots & \qquad\quad \nonumber  \\
\label{eq:drawD}
z_D & \sim  q(z_D|z_1,z_2,\ldots,z_{D-1}).
\end{align}
This basic strategy can also be used to sample from distributions whose domains are Cartesian products of different one-dimensional spaces.

In what follows, we first introduce a ``core primitive'' density,
    and we show how to sample efficiently from its distribution.
    We then show how the sampling procedure in \Crefrange{eq:draw1}{eq:drawD}
    reduces to sampling from this core primitive; a key component of this procedure
    is the property of orthogonality, which helps facilitate the efficient computation of
    marginal distributions.

\subsubsection*{Core primitive}

First we describe the core primitive that we will use for each of the draws in
\Crefrange{eq:draw1}{eq:drawD}.
To begin, we observe the following: if $S$ is any positive semidefinite matrix with $\text{trace}(S)\!=\!1$, then
\begin{equation}
\rho(\xi) = \sum_{k,\ell=1}^K S_{k\ell} \phi_k(\xi)\phi_\ell(\xi),
\label{eq:psd-density}
\end{equation}
defines a normalized density over $\mathbb{R}$. In particular, since $S\succeq 0$, it follows that $\rho(\xi)\!\geq\! 0$ for all $\xi\!\in\!\mathbb{R}$, and since $\text{trace}(S)\!=\!1$, it follows that
\begin{equation}
\int_{-\infty}^\infty\!\! \rho(\xi)\, d\xi
  = \sum_{k,\ell=1}^K S_{k\ell} \int_{-\infty}^\infty\!\! \phi_k(\xi)\phi_\ell(\xi)\, d\xi
  = \sum_{k,\ell=1}^K S_{k\ell} \delta_{kl}
  = \text{trace}(S)
  = 1.
\end{equation}
The core primitive that we need is an efficient procedure to sample
from a normalized density of this form. We will see later that all of the densities
in
\Crefrange{eq:draw1}{eq:drawD}
can be expressed in this~form.

\subsubsection*{Inverse transform sampling}
Since the density in
\Cref{eq:psd-density}
is one-dimensional, we can obtain the draw we need by inverse transform sampling.
In particular, let $\C(\xi)$ denote the cumulative distribution function (CDF)
associated with \Cref{eq:psd-density}, which is given by
\begin{equation}
\C(\xi) = \int_{-\infty}^\xi\!\! \rho(z)\,dz,
\label{eq:CDF}
\end{equation}
and let $\C^{-1}(\xi)$ denote the inverse CDF.
Then at least in principle, we can draw a sample from $\rho$ by the two-step procedure
\begin{align}
    u & \sim \text{Uniform}[0,1], \\
    \xi & = \C^{-1}(u). \label{eq:invCDF}
\end{align}
Next we consider how to implement this procedure efficiently in practice,
and in particular, how to calculate the definite integral for the CDF in \Cref{eq:CDF}.
As shorthand, we define the doubly-indexed set of real-valued functions
\begin{equation}
    \Phi_{k\ell}(\xi) = \int_{-\infty}^\xi \phi_k(z)\phi_\ell(z)\, dz.
\end{equation}
It follows from orthogonality that $\Phi_{kl}(+\infty) = \delta_{kl}$ and from the Cauchy-Schwartz inequality that $|\Phi_{k\ell}(\xi)|\leq 1$ for all $\xi\in\mathbb{R}$. Our interest in these functions stems from the observation that
\begin{equation}
\C(\xi) = \sum_{k,\ell=1}^K S_{k\ell} \Phi_{kl}(\xi) = \text{trace}[S\Phi(\xi)],
\label{eq:CDF-trace}
\end{equation}
so that if we have already computed the functions $\Phi_{k\ell}(\xi)$,
then we can use \Cref{eq:CDF-trace} to compute the CDF whose inverse we need in
\Cref{eq:invCDF}.
In practice, we can use numerical quadrature to pre-compute $\Phi_{k\ell}(\xi)$
for many values along the real line and then solve \Cref{eq:invCDF} quickly by interpolation;
that is, given $u$, we find $\xi$ satisfying $\text{trace}[S\Phi(\xi)]=u$.
The result is an unbiased sample drawn from the density $\rho(\xi)$ in \Cref{eq:psd-density}.

\subsubsection*{Sequential sampling}
Finally we show that each draw in
\Crefrange{eq:draw1}{eq:drawD}
reduces to the problem described above.
As in \Cref{sec:orth}, we work out the steps specifically for an example in $D\!=\!3$,
where we must draw the samples $z_1\sim q(z_1)$, $z_2\sim q(z_2|z_1)$ and
$z_3\sim q(z_3|z_1,z_2)$.
This example illustrates all the ideas needed for the general case but with a
minimum of indices.

Consider the joint distribution given by
\begin{equation}
q(z_1,z_2,z_3) = \left(\sum_{i=1}^{K_1}\sum_{j=1}^{K_2}\sum_{k=1}^{K_3} \beta_{ijk}\, \phi_i(z_1)\phi_j(z_2)\phi_k(z_3)\right)^2\quad\mbox{where}\quad \sum_{ijk}\beta^2_{ijk} = 1.
\label{eq:3d-redux}
\end{equation}
From this joint distribution, we can compute marginal distributions by
integrating out subsets of variables, and each integration over $\mathbb{R}$
gives rise to a contraction of indices, as in \Cref{eq:marginal}, due to the
property of orthogonality.
{In particular, expanding the square in
\Cref{eq:3d-redux}, we can write this joint distribution as
\begin{equation}
q(z_1,z_2,z_3) =
    \sum_{k,k'=1}^{K_3}
    \left[\sum_{i,i'=1}^{K_1}
    \sum_{j,j'=1}^{K_2}
\beta_{ijk}\,
\beta_{i'j'k'}\,
    \phi_i(z_1) \phi_{i'}(z_1) \phi_j(z_2) \phi_{j'}(z_2)
    \right] \phi_k(z_3) \phi_{k'}(z_3),
\label{eq:3d-redux2}
\end{equation}
and we can then contract the index $k'$ when integrating over $z_3$, since $\int \phi_k(z_3) \phi_{k'}(z_3) dz_3 = \delta_{kk'}$.
}

In this way we find that the marginal distributions are
\begin{align}
q(z_1,z_2) &= \sum_{j,j'=1}^{K_2} \left[\sum_{i,i'=1}^{K_1}\sum_{k=1}^{K_3} \beta_{ijk}\beta_{i'j'k} \phi_i(z_1)\phi_{i'}(z_1)\right]\phi_j(z_2)\phi_{j'}(z_2), \label{eq:marg2} \\
q(z_1) &= \sum_{i,i'=1}^{K_1} \left[\sum_{j=1}^{K_2}\sum_{k=1}^{K_3} \beta_{ijk}\beta_{i'jk}\right] \phi_i(z_1)\phi_{i'}(z_1).\label{eq:marg1}
\end{align}

Now note from the brackets in \Cref{eq:marg1} that this marginal distribution
is already in the quadratic form of \Cref{eq:psd-density} with coefficients
\begin{equation}
    S^{(1)}_{ii'} = \sum_{j=1}^{K_2}\sum_{k=1}^{K_3} \beta_{ijk}\beta_{i'jk}.
\end{equation}
From this first quadratic form, we can therefore use inverse transform sampling
to obtain a draw $z_1 \sim q(z_1)$.

Next we consider how to sample from the conditional
$q(z_2|z_1) = q(z_1,z_2)/q(z_1)$.
Again, from the brackets in \Cref{eq:marg2}, we see that this
conditional distribution is also in the quadratic form of \Cref{eq:psd-density}
with coefficients
\begin{equation}
    S_{jj'}^{(2)} = \frac{\sum_{i,i'=1}^{K_1}\sum_{k=1}^{K_3} \beta_{ijk}\beta_{i'j'k}\phi_i(z_1)\phi_{i'}(z_1)}{q(z_1)}.
\end{equation}
From this second quadratic form, we can therefore use inverse transform sampling
to obtain a draw $z_2 \sim q(z_2|z_1)$.
Finally, we consider how to sample from $q(z_3|z_1,z_2) = q(z_1,z_2,z_3)/q(z_1,z_2)$.
From \Cref{eq:3d-redux2}, we see that this conditional distribution is also in
the quadratic form of \Cref{eq:psd-density} with coefficients
\begin{equation}
    S_{kk'}^{(3)} = \frac{\sum_{i,i'=1}^{K_1}\sum_{j,j'=1}^{K_2}\beta_{ijk}\beta_{i'j'k'}\phi_i(z_1)\phi_{i'}(z_1)\phi_j(z_2)\phi_{j'}(z_2)}
    {q(z_1,z_2)}
    \label{eq:sampling3}
\end{equation}
From this third quadratic form, we can therefore use inverse transform sampling to obtain a draw $z_3\sim q(z_3|z_1,z_2)$.
Finally, from the sums in \Cref{eq:sampling3}, we see that the overall cost of
this procedure is $\O(K_1^2 K_2^2 K_3^2)$, or quadratic in the total number of
basis functions.




\section{Calculation of moments}
\label{app:moments}

In this appendix we show how to calculate the low-order moments of a density
constructed from the Cartesian product of orthogonal function expansions.
In particular, we assume that the density is over~$\mathbb{R}^D$ and of the form
\begin{equation}
    q(z_1,z_2,\ldots,z_D) = \left(\sum_{k_1=1}^{K_1} \cdots \sum_{k_D=1}^{K_D} \alpha_{k_1 k_2 \ldots k_D}\phi_{k_1}(z_1)\phi_{k_2}(z_2)\cdots\phi_{k_D}(z_D)\right)^2,
    \label{eq:joint}
\end{equation}
where $\{\phi_{k}(\cdot)\}_{k=1}^\infty$ are orthogonal functions on $\mathbb{R}$ and where the coefficients are properly normalized so that the density integrates to one. For such a density, we show that the calculation of first and second-order moments boils down to evaluating \textit{one-dimensional} integrals of the form
\begin{align}
\mu_{ij} &= \int_{-\infty}^\infty\! \phi_i(z)\phi_j(z)\, z\, dz,  \label{eq:momint1} \\
\nu_{ij} &= \int_{-\infty}^\infty\! \phi_i(z)\phi_j(z)\, z^2\, dz. \label{eq:momint2}
\end{align}
We also show how to evaluate these integrals specifically for the orthogonal family of weighted Hermite polynomials.

First we consider how to calculate moments such as $\mathbb{E}_q[z_d^p]$, where $p\in\{1,2\}$, and without loss of generality we focus on calculating $\mathbb{E}_q[z_1^p]$.
We start from the joint distribution in \Cref{eq:joint}
and proceed by marginalizing over the variables $(z_2,z_3,\ldots,z_D)$.
Exploiting orthogonality, we find that
\begin{align}
  \mathbb{E}_q[z_1^p]
    &= \int\! q(z_1,z_2,\ldots,z_D)\, z_1^p\, dz_1\, dz_2\, \ldots dz_D, \\
    &= \int\! \left(\sum_{k_1=1}^{K_1} \cdots \sum_{k_D=1}^{K_D} \alpha_{k_1 k_2 \ldots k_D}\phi_{k_1}(z_1)\phi_{k_2}(z_2)\cdots\phi_{k_D}(z_D)\right)^2\! z_1^p\, dz_1\, dz_2\, \ldots dz_D, \\
   &= \sum_{k_1,k'_1=1}^{K_1} \left[\sum_{k_2=1}^{K_2} \cdots \sum_{k_D=1}^{K_D} \alpha_{k_1 k_2 \ldots k_D} \alpha_{k'_1 k_2 \ldots k_D}\right] \int\! \phi_{k_1}(z_1)\phi_{k'_1}(z_1)\, z_1^p\, dz_1. \label{eq:marginal1}
\end{align}
We can rewrite this expression more compactly as a quadratic form over integrals
of the form in \Crefrange{eq:momint1}{eq:momint2}.
To this end, we define the coefficients
\begin{equation}
    A_{ij} = \sum_{k_2=1}^{K_2}\cdots\sum_{k_D=1}^{K_D} \alpha_{i k_2 \ldots k_D} \alpha_{j k_2
    \ldots k_D},
    \label{eq:Aij}
\end{equation}
which simply encapsulate the bracketed term in \Cref{eq:marginal1}.
Note that there are $K_1^2$ of these coefficients, each of which can be
computed in $\O(K_2 K_3\ldots K_D)$.
With this shorthand, we can write
\begin{align}
    \mathbb{E}_q[z_1] &= \sum_{i,j=1}^{K_1} A_{ij} \mu_{ij}, \label{eq:Amu}\\
    \mathbb{E}_q[z_1^2] & = \sum_{i,j=1}^{K_1} A_{ij} \nu_{ij}, \label{eq:Anu}
\end{align}
where $\mu_{ij}$ and $\nu_{ij}$ are the integrals defined in
\Crefrange{eq:momint1}{eq:momint2}.
Thus the problem has been reduced to a weighted sum of one-dimensional integrals.

A similar calculation gives the result we need for correlations. Again, without loss of generality, we focus on calculating $\mathbb{E}_q[z_1 z_2]$.
Analogous to \Cref{eq:Aij}, we define the tensor of coefficients
\begin{equation}
B_{ijk\ell} = \sum_{k_3=1}^{K_3}\cdots\sum_{k_D=1}^{K_D} \alpha_{i k k_3 \ldots k_D} \alpha_{j \ell k_3 \ldots k_D},
\end{equation}
which arises from marginalizing over the variables $(z_3,z_4,\ldots,z_D)$.
There are $K_1^2 K_2^2$ of these coefficients, each of which can be computed
in $\O(K_3 K_4\ldots K_D)$. With this shorthand, we can write
\begin{equation}
    \mathbb{E}_q[z_1 z_2] = \sum_{i,j=1}^{K_1} \sum_{k,\ell=1}^{K_2} B_{ijk\ell}\mu_{ij}\mu_{k\ell}.
    \label{eq:Bmumu}
\end{equation}
where $\mu_{ij}$ is again the integral defined in \Cref{eq:momint1}).
Thus the problem has been reduced to a weighted sum of (the product of)
one-dimensional integrals.

Finally, we show how to evaluate the integrals in \Crefrange{eq:momint1}{eq:momint2}
for the specific case of orthogonal function expansions with weighted Hermite polynomials;
{similar computations apply in the case of Legendre polynomials.}
Recall in this case that
\begin{equation}
\phi_{k+1}(z) = \left(\sqrt{2\pi}k!\right)^{-\frac{1}{2}}\left(e^{-\frac{1}{2}z^2}\right)^{\frac{1}{2}}\,\text{H}_{k}(z),
\label{eq:phiH-def}
\end{equation}
where $\text{H}_k(z)$ are the \emph{probabilist's} Hermite polynomials given by
\begin{equation}
\text{H}_k(z) = (-1)^k e^{\frac{z^2}{2}} \frac{d^k}{dz^k}\left[e^{-\frac{z^2}{2}}\right].
\end{equation}
To evaluate the integrals for this particular family, we can exploit the following recursions that are satisfied by Hermite polynomials:
\begin{align}
  H_{k+1}(z) &= zH_k(z) - H'_k(z), \label{eq:Hrecur1} \\
  H'_k(z) &= kH_{k-1}(z). \label{eq:Hrecur2}
\end{align}
Eliminating the derivatives $H'_k(z)$ in \Crefrange{eq:Hrecur1}{eq:Hrecur2},
we see that $zH_k(z) = H_{k+1}(z) + kH_{k-1}(z)$.
We can then substitute \Cref{eq:phiH-def} to obtain a recursion
for the orthogonal basis functions themselves:
\begin{equation}
z\phi_k(z) = \sqrt{k}\phi_{k+1}(z) + \sqrt{k\!-\!1}\phi_{k-1}(z).
\label{eq:phi-recur}
\end{equation}
With the above recursion, we can now read off these integrals from the property of orthogonality.
For example, starting from \Cref{eq:momint1}, we find that
\begin{align}
\mu_{ij}
  &= \int_{-\infty}^\infty\! \phi_i(z)\phi_j(z)\, z\, dz, \\
  &= \int_{-\infty}^\infty\! \phi_i(z) \left[\sqrt{j}\phi_{j+1}(z) + \sqrt{j\!-\!1}\phi_{j-1}(z)\right]\, dz, \\
  &= \delta_{i,j+1}\sqrt{j} + \delta_{i,j-1}\sqrt{i}, \label{eq:mom1ij}
\end{align}
where $\delta_{ij}$ is the Kronecker delta function.
Next we consider the integral in \Cref{eq:momint2},
which involves a power of $z^2$ in the integrand. In this case we can make repeated use of the recursion:
\begin{align}
\nu_{ij}
  &= \int_{-\infty}^\infty\! \phi_i(z)\phi_j(z)\, z^2\, dz, \\
  &= \int_{-\infty}^\infty\! \left[\sqrt{i}\phi_{i+1}(z) + \sqrt{i\!-\!1}\phi_{i-1}(z)\right]\, \left[\sqrt{j}\phi_{j+1}(z) + \sqrt{j\!-\!1}\phi_{j-1}(z)\right]\, dz, \\
  &= \delta_{ij}\left[\sqrt{ij}+\sqrt{(i\!-\!1)(j\!-\!1)}\right]\,
      +\, \delta_{i-1,j+1}\sqrt{j(j\!+\!1)}\,
      +\, \delta_{j-1,i+1}\sqrt{i(i\!+\!1)}.
      \label{eq:mom2ij}
\end{align}
Note that the matrices in \Cref{eq:mom1ij} and \Cref{eq:mom2ij}
can be computed for whatever size is required by the orthogonal basis function
expansion in \Cref{eq:joint}.
Once these matrices are computed, it is a simple matter of
substitution\footnote{With further bookkeeping, one can also exploit the
\textit{sparsity} of $\mu_{ij}$ and $\nu_{ij}$ to derive more efficient
calculations of these moments.}
to compute the moments $\mathbb{E}_q[z_1]$, $\mathbb{E}_q[z_1^2]$, and
$\mathbb{E}_q[z_1 z_2]$ from \Crefrange{eq:Amu}{eq:Anu} and \Cref{eq:Bmumu}.
Finally, we can compute other low-order moments
(such as $\mathbb{E}_q[z_5]$ or $\mathbb{E}_q[z_3 z_7]$) by an appropriate
permutation of indices.




\section{Eigenvalue problem}
\label{app:eigen}

In this appendix we show in detail how the optimization for EigenVI reduces to a minimum eigenvalue problem. In particular we prove the following.

\begin{lemma}\label{lem:obj-min-eigen}
Let $\{\phi_k(z)\}_{k=1}^\infty$ be an orthogonal function expansion, and let $q\in\mathcal{Q}_K$ be the variational approximation parameterized by
     \begin{align}
        q(z) = \left[\sum_{k=1}^K \alpha_k \phi_k(z)\right]^2,
    \end{align}
where the weights satisfy $\sum_{k=1}^K \alpha_k^2=1$, thus ensuring that the distribution is normalized.
Suppose furthermore that $q$ is chosen to minimize the empirical estimate of the Fisher divergence given, as in eq.~(\ref{eq-empirical-divergence}), by
\begin{displaymath}
 \widehat\D_{\pi}(q, p) =
    \sum_{b=1}^B \frac{q(z^b)}{\pi(z^b)} \, \big\|\nabla \log q(z^b) - \nabla\log p(z^b)\big\|^2.
\end{displaymath}
Then the optimal variational approximation $q$ in this family
    can be computed by solving the minimum eigenvalue problem
    \begin{align}
        \label{eq:eigenVI-solution-ap}
\min_{q\in\Q_K}\left[\widehat{\D}_{\pi}(q,p)\right] = \min_{\|\alpha\|=1}
      \alpha^\top M \alpha
        =: \lambda_{\text{min}}(M),
      \end{align}
      where $M$ is given in~\Cref{eq:M} and  $\alpha =[\alpha_1, \ldots, \alpha_K]\in \R^K$.
    The optimal weights $\alpha$ are given (up to an arbitrary sign) by the corresponding eigenvector of this minimal eigenvalue.
\end{lemma}
\begin{proof}
    The scores of $q$ in this variational family are given by
\[\nabla\log q(z^b) = \frac{2\sum_k \alpha_k \nabla \phi_k(z^b)}{\sum_k \alpha_k \phi_k(z^b)}. \]
Substituting the above into the empirical divergence, we find that
\begin{align*}
    \widehat\D_{\pi}(q, p) &=   \sum_{b=1}^B \frac{q(z^b)}{\pi(z^b)} \, \big\|\nabla \log q(z^b) - \nabla\log p(z^b)\big\|^2\\
    &=  \sum_{b=1}^B \frac{\big(\sum_{k} \alpha_k \phi_k(z^b)\big)^2}{\pi(z^b)} \, \left\|\frac{2\sum_k \alpha_k \nabla \phi_k(z^b)}{\sum_k \alpha_k \phi_k(z^b)} - \nabla\log p(z^b)\right\|^2 \\
    &=  \sum_{b=1}^B \frac{1}{\pi(z^b)} \, \left\|2\sum_k \alpha_k \nabla \phi_k(z^b) - \bigg[\sum_{k}
    \alpha_k \phi_k(z^b)\bigg]\nabla\log p(z^b)\right\|^2\\
    &= \sum_{b=1}^B \frac{1}{\pi(z^b)} \, \left\|\sum_k \alpha_k \left[ 2\nabla \phi_k(z^b) -
\phi_k(z^b)\nabla\log p(z^b)\right]\right\|^2 \\
    &= \alpha^\top M \alpha,
\end{align*}
where $M$ is given in~\eqref{eq:M} and  $\alpha =[\alpha_1, \ldots, \alpha_K]\in \R^K$. Thus the optimal weights $\alpha$ are found by minimizing the quadratic form $\alpha^\top M\alpha$ subject to the constraint $\alpha^\top\alpha=1$. Equivalently, a solution can be found by minimizing the Rayleigh quotient
\begin{equation}
\argmin_v \frac{v^\top M v}{v^\top v}
\end{equation}
and setting $\alpha=v/\|v\|$. It then follows from the Rayleigh-Ritz theorem~\citep{courant1924methoden} for symmetric matrices that $\alpha$ is the eigenvector corresponding to the minimal eigenvalue of $M$, and this proves the lemma.

\end{proof}




\section{Practical considerations of EigenVI}
\label{sec:discussion:eigenvi}

\subsection{EigenVI vs gradient-based BBVI}

Recall that EigenVI has two hyperparameters: the number of basis functions
$K$ and the number of importance samples $B$.
We note there is  an important difference between these two hyperparameters
and the learning rate in ADVI and other gradient-based methods.
Here, as we use more basis functions and more samples, the resulting fit
is a better approximation. So, we can increase the number of basis functions
and importance samples until a budget is reached, or until the resulting variational approximation is a sufficient fit.
On the other hand, tuning the learning rate in gradient-based optimization
is much more sensitive because it cannot be too large or too small. If it is too large, ADVI may diverge. If the learning rate is too small, it may take too long to converge in which case it may exceed computational budgets.

Another fundamental difference in setting the number of basis functions as compared to the
learning rate or batch size of gradient based optimization is that once we have evaluated the score of the target distribution for the samples, these same samples can be reused for solving the eigenvalue problem
with any choice of the number of basis functions, as these tasks are independent.
By contrast, in iterative BBVI, the optimization problem needs to be re-solved
for every choice of hyperparameters,
and the samples from different runs cannot be mixed together.

Furthermore, solving
the eigenvalue problem is fast, and scores can be computed in parallel.
In our implementation, we use off-the-shelf eigenvalue solvers, such as ARPACK~\cite{arpack1998} or
Julia's eigenvalue decomposition function, \texttt{eigen}.
In many problems with complicated targets, the main cost comes from gradient evaluation and not the eigenvalue solver.

\subsection{Choosing the number of samples $B$}

Intuitively, if the target $p$ is in the variational family $\Q$ (i.e., it can be represented
using an order-$K$ expansion), then we should choose
the number of samples $B$ to roughly equal the number of basis function $K$.
If $p$ is very different from $\Q$, we need more samples, and in our experiments,
we use a multiple of the number of basis functions (say of order 10).  As discussed before, once we have evaluated a set of scores, these can be reused to fit a larger number of basis functions.

\section{Additional experiments and details}

\subsection{Computational resources}

The experiments were run on a Linux workstation with
a 32-core Intel(R) Xeon(R) w5-3435X processor
and with 503 GB of memory.
Experiments were run on CPU.
In the sinh-arcsinh and posteriordb experiments,
computations to construct the matrix $M$ were parallelized
over 28 threads.

\subsection{2D synthetic targets}
\label{ssec-2dsynthetic}

\begin{figure}[t]
    \centering
    \includegraphics[scale=0.42]{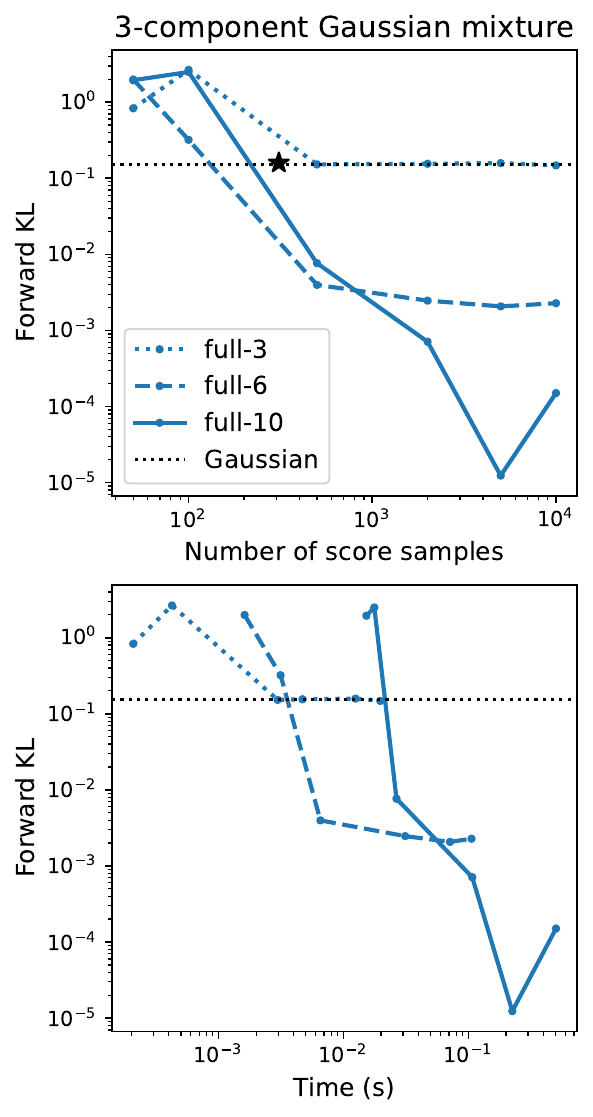}
    \includegraphics[scale=0.42]{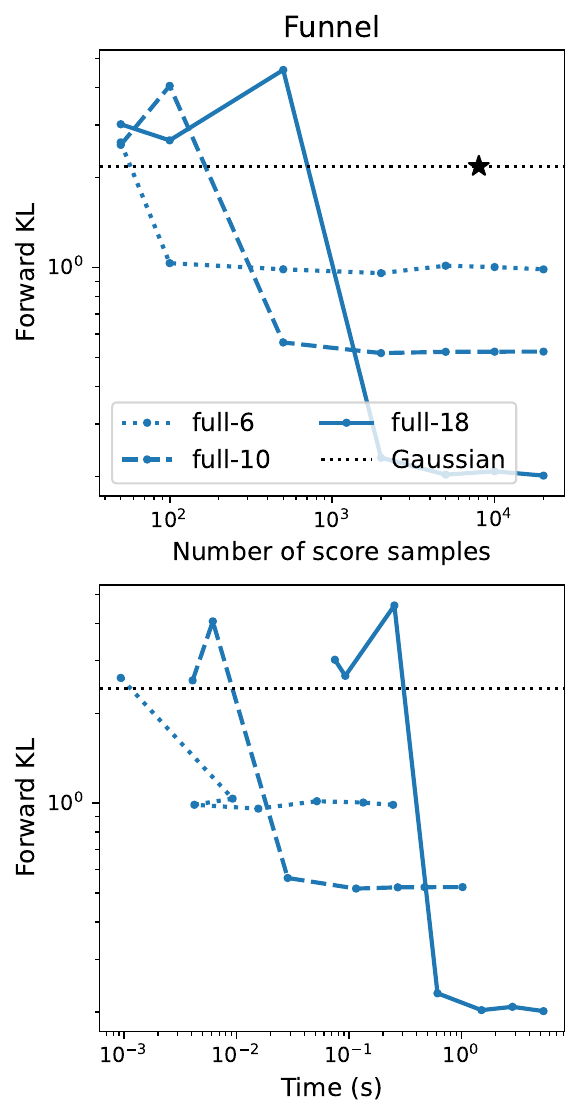}
    \includegraphics[scale=0.42]{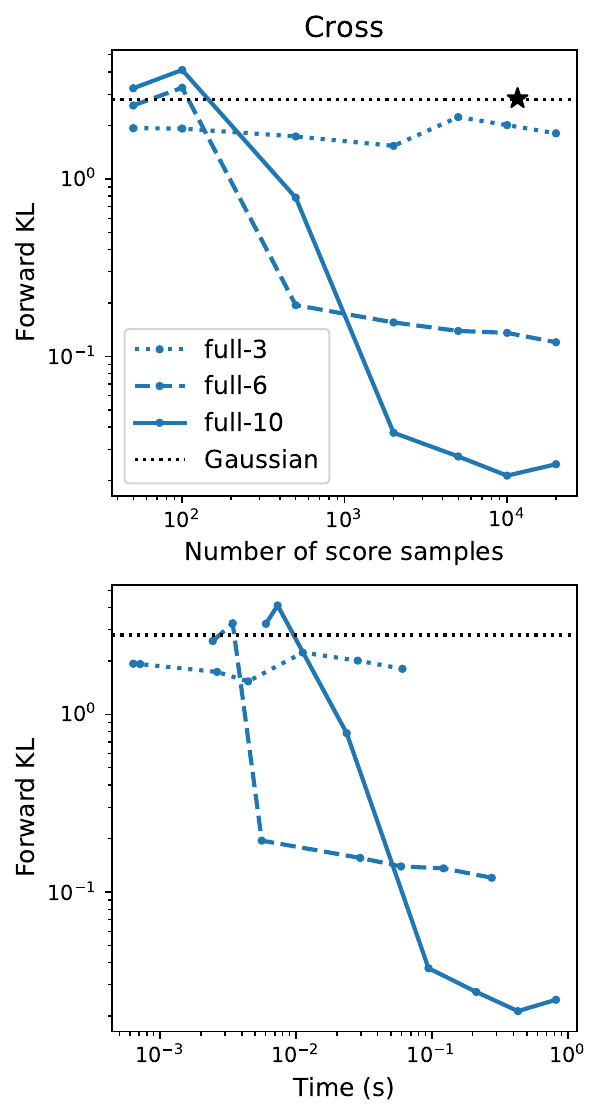}
    \caption{We compare the number of score evaluations wallclock vs
FKL divergence for the target distributions in
\Cref{fig:2dtargets}: the Gaussian mixture (column 1),
the funnel (column 2), and
the cross (column 3) distributions.
The $K$ used for EigenVI is reported in each figure legend,
where $K=K_1 K_2$.
The black star denotes the number of gradient evaluations for the Gaussian method.
    }
\label{fig:2dtargets-metrics}
\end{figure}

We considered the following synthetic 2D targets:

\begin{itemize}

    \item \textbf{3-component Gaussian mixture:}
    \begin{align*}
    p(z) =
    0.4 \N(z \given [-1,1]^\top,\Sigma)
    +
    0.3 \N(z \given  [1.1,1.1]^\top,0.5I)
    +
    0.3 \N(z \given [-1,-1]^\top,0.5I),
\end{align*}
where we define
$\Sigma =
\begin{bmatrix}
    2 & 0.1 \\
    0.1 & 2 \\
\end{bmatrix}$.

\item \textbf{Funnel distribution with $\sigma^2=1.2$:}
\begin{align*}
    p(z) = \N(z_1 \given 0, \sigma^2) \N(z_2 \given 0, \exp(z_1/2)).
\end{align*}

\item \textbf{ Cross distribution:}
\begin{align*}
    p(z) =
    \tfrac{1}{4} \N(z \given [0,2]^\top,\Sigma_1)
    +
     \tfrac{1}{4} \N(z \given [-2,0]^\top,\Sigma_2)
    +
     \tfrac{1}{4} \N(z \given  [2,0]^\top,\Sigma_2)
    +
     \tfrac{1}{4} \N(z \given  [0,-2]^\top,\Sigma_1),
\end{align*}
where
$\Sigma_1 = \begin{bmatrix}
0.15^{0.9} & 0 \\ 0 & 1
\end{bmatrix}$
and
$\Sigma_2 = \begin{bmatrix}
    1 & 0 \\
0 & 0.15^{0.9}
\end{bmatrix}$.

\end{itemize}

These experiments were conducted without standardization with a Gaussian VI estimate.
The EigenVI proposal distribution $\pi$ used was a $\text{uniform}([-9,9])$ distribution.

In  \Cref{fig:5dtargetdensity}, we run EigenVI for increasing numbers of importance samples $B$
and report the resulting forward KL divergence.
The blue curves denote variational families with different $K_1=K_2=K$ values used,
i.e., 3, 6, and 10 (resulting in a total
number of basis functions of $3^2$, $6^2$, and $10^2$).
In the bottom row of the plot, we also show wall clock timings (computed without parallelization)
to show how the cost grows with the increase in the number of basis functions and importance samples.
The horizontal dotted line denotes the result from batch and match VI,
which fits a Gaussian via score matching; here a batch size of 16 was used
and a learning rate of $\lambda_t = \tfrac{BD}{t+1}$.

The black star denotes the number of score evaluations used by the Gaussian
VI method.

\subsection{Sinh-arcsinh targets}
\label{ssec-sinh}

The sinh-arcsinh normal distribution \citep{jones2009sinh,jones2019sinh}
has parameters $s \in \reals^D, \tau \in \reals_+^D, \Sigma \in \mathbb{S}_{++}$;
it is induced by transforming a
Gaussian $Z_0 \sim \N(0, \Sigma)$ to $Z = \mathcal{S}_{s,\tau}(Z_0)$,
where
\begin{align}
\mathcal{S}_{s,\tau}(z) :=
    [S_{s_1,\tau_1}(z_1), \ldots, S_{s_D,\tau_D}(z_D)]^\top,
    \quad
S_{s_d,\tau_d}(z_d) := \sinh\left(\tfrac{1}{\tau_d} \sinh^{-1}(z_d) + \tfrac{s_d}{\tau_d}\right).
\end{align}
Here $s_d$ controls the amount of skew in the $d$th dimension,
and $\tau_d$ controls the tail weight in that dimension.
When $s_d\!=\!{0}$ and $\tau_d\!=\!{1}$ in all dimensions $d$, the distribution is Gaussian. 

The sinh-arcsinh normal distribution has the following density:
\begin{align}
p(z; s, \tau, \Sigma)
    = [(2\pi)^D |\Sigma|]^{-\frac{1}{2}}
    \prod_{d=1}^D
    \left\{
        (1 + z_d^2)^{-\frac{1}{2}}
\tau_d \, C_{s_d, \tau_d}(z_d)
    \right\}
\exp\left(-\frac{1}{2} S_{s,\tau}(z)^\top \Sigma^{-1} S_{s,\tau}\right),
\end{align}
where we define the functions
\begin{align}
    C_{s_d,\tau_d}(z_d) := (1 + S_{s_d,\tau_d}^2(z))^{\frac{1}{2}},
\end{align}
and
\begin{align}
    \quad
    S_{s_d,\tau_d}(z_d) := \sinh(\tau_d \sinh^{-1}(z_d) - s_d),
    \quad
    S_{s,\tau}(z) =
    [S_{s_1,\tau_1}(z_1), \ldots, S_{s_D,\tau_D}(z_D)]^\top.
\end{align}

We constructed 3 targets in 2 dimensions and 3 targets in 5 dimensions,
each with varying amounts of non-Gaussianity.
The details of each target are below.
In all experiments, EigenVI was applied with standardization,
where a Gaussian was fit using batch and match VI with a batch size of 16
and a learning rate $\lambda_t = \tfrac{BD}{t+1}$.

For all experiments, we used a proposal distribution $\pi$
that was uniform on $[-5,5]^2$.

\begin{figure}[t]
    \centering
    \includegraphics[scale=0.28]{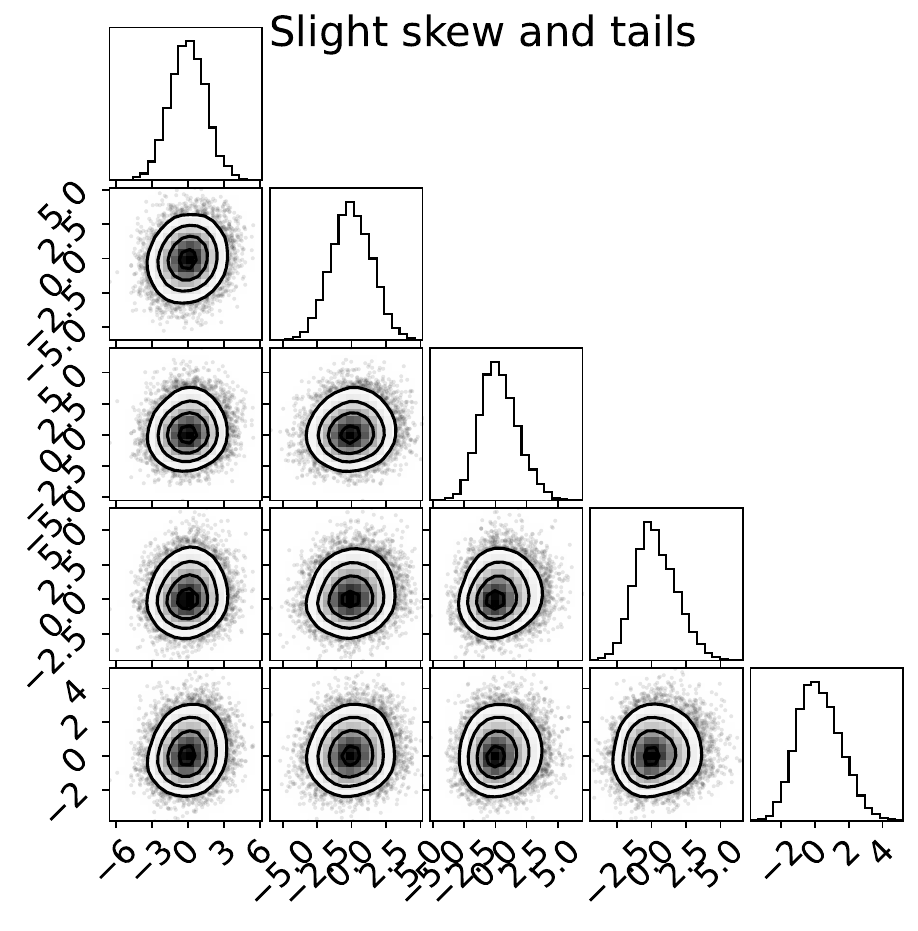}
    \includegraphics[scale=0.28]{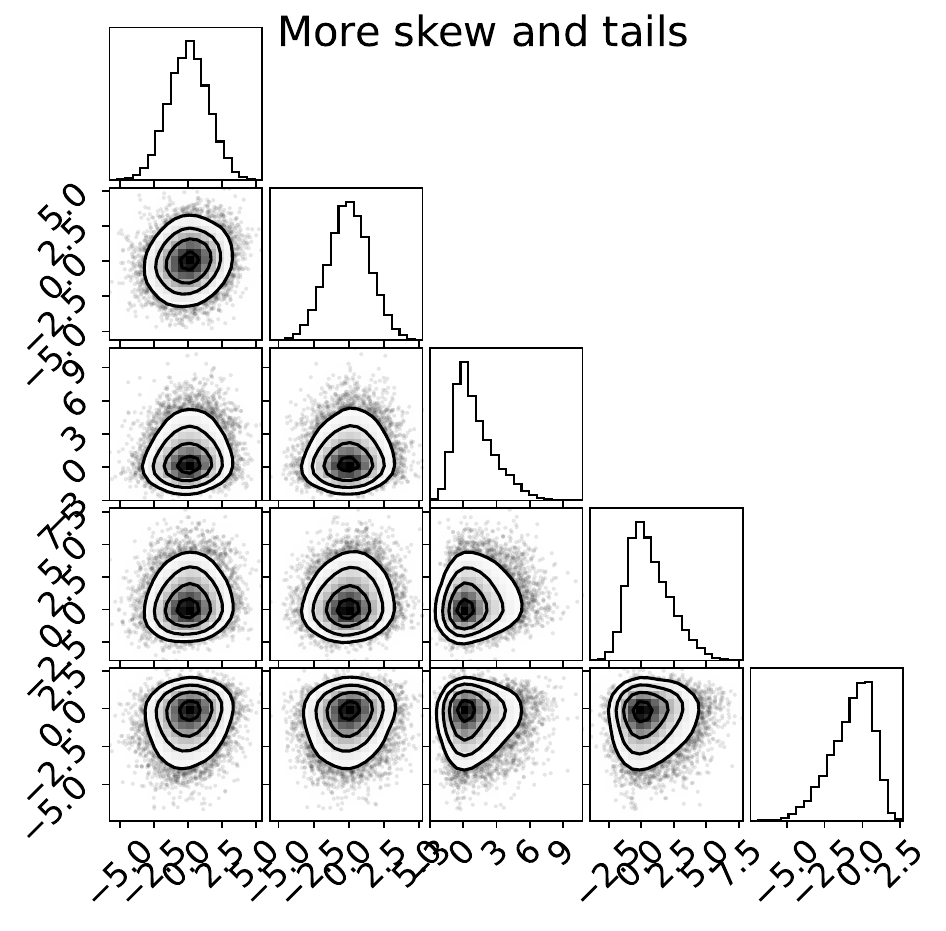}
    \includegraphics[scale=0.28]{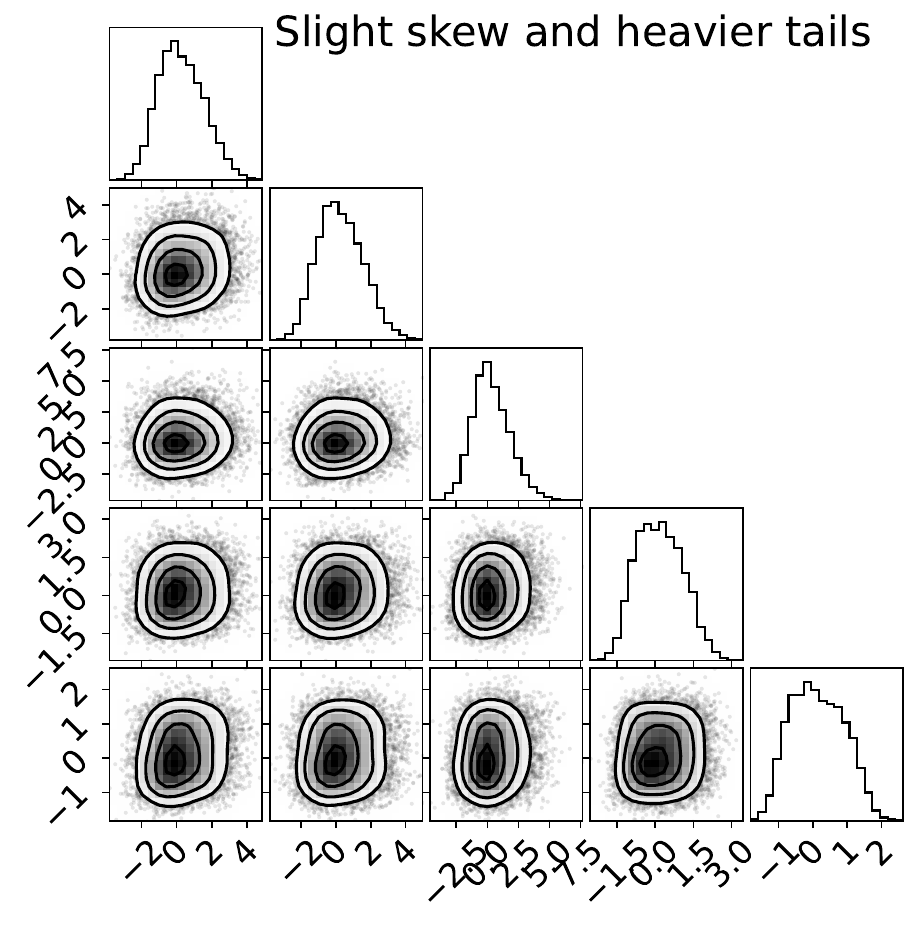}
    \includegraphics[scale=0.28]{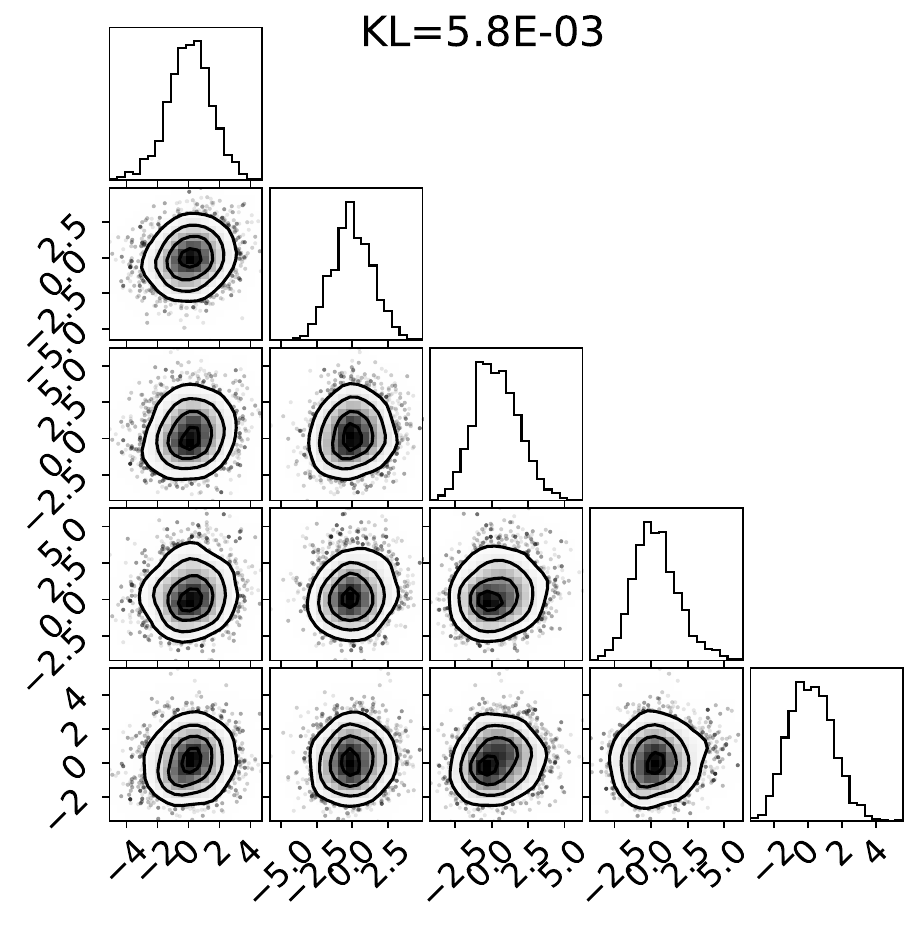}
    \includegraphics[scale=0.28]{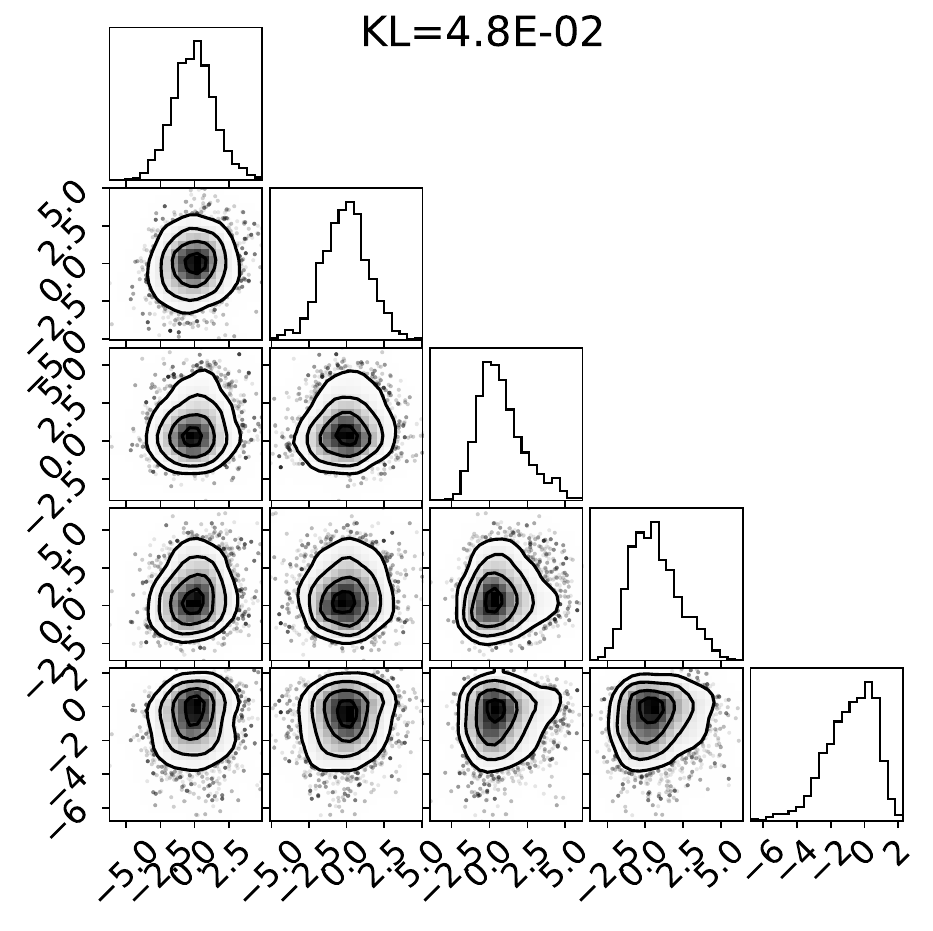}
    \includegraphics[scale=0.28]{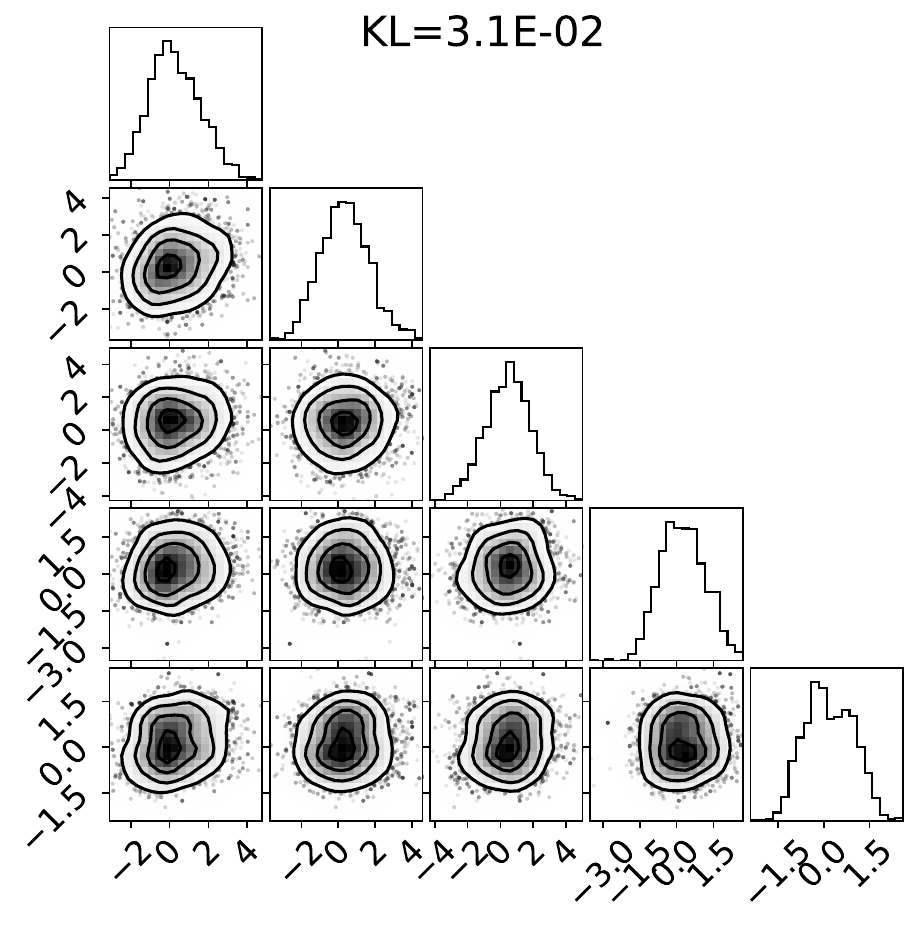}
\caption{Targets (top) for the 5D sinh-arcsinh normal distribution example
and EigenVI fits (bottom) with the KL divergence in the figure title.}
\label{fig:5dtargetdensity}
\end{figure}

\paragraph{2D sinh-arcsinh normal experiment}
For $D=2$ (\Cref{subfig:sinh2D}), we consider the
\emph{slight skew and tails target} with parameters
$s=[0.2,0.2], \tau = [1.1,1.1]$,
the \emph{more skew and tails target} with
$s=[0.2,0.5], \tau = [1.1,1.1]$,
and \emph{the slight skew and heavier tails} with
$s=[0.2,0.2], \tau = [1.4,1.1]$.
Note that $s=[0,0], \tau=[1,1]$ recovers the multivariate Gaussian.
These three target are visualized in
\Cref{fig:sinh2dtargets}.

\paragraph{5D sinh-archsinh normal experiment}

We constructed three targets $P_1$ (slight skew and tails), $P_2$ (more skew and tails),
and $P_3$ (slight skew and heavier tails) each with
\begin{align}
    \Sigma =
\begin{bmatrix}
    2.2 & 0.3 & 0   & 0   & 0.3 \\
    0.3 & 2.2 & 0   & 0   & 0 \\
    0   & 0   & 2.2 & 0.3 & 0 \\
    0   & 0   & 0.3 & 2.2 & 0 \\
    0.3 & 0   & 0   & 0   & 2.2 \\
\end{bmatrix}.
\end{align}
The skew and tail weight parameters used were:
$s_1 = [0.,0.,0.2,0.2,0.2]; \tau_1 = [1., 1., 1., 1., 1.1]$,
$s_2 = [0.0,0.0,0.6,0.4,-0.5]; \tau_2 = [1., 1., 1., 1., 1.1]$,
and
$s_3 = [0.2,0.2,0.2,0.2,0.2]; \tau_3 = [1.1, 1.1, 1., 1.4, 1.6]$.
See \Cref{fig:5dtargetdensity} for a visualization of the marginals of each target distribution.
In the second row, we show examples of resulting EigenVI fit (visualized using samples from $q$)
from $B=20{,}000$ and $K=10$.

\subsection{Posteriordb experiments}
\label{ssec:pdb}

We consider 8 real data targets from \texttt{posteriordb}, a suite of benchmark
Bayesian models for real data problems.
In \Cref{table:posteriordb}, we summarize the models considered in the study.
These target distributions are non-Gaussian, typically with some skew or different tails.
To access the log target probability and their gradients, we used the
BridgeStan library \citep{roualdes2023bridgestan}, which by default transforms the target
to be supported on $\reals^D$.

For all experiments, we fixed the number of importance samples to be
$B=40{,}000$; to construct the EigenVI matrix $M$, the computations were parallelized
over the samples.
These experiments were repeated over 5 random seeds, and we report the mean and standard errors in
\Cref{fig:posteriordb}; for lower dimensions, there was little variation between runs.

The target distributions were standardized using a Gaussian fit from score matching
before applying EigenVI.
In most cases, the proposal distribution $\pi$ was chosen to be uniform over $[-6,6]^D$.
For the models \texttt{8-schools},
which has a longer tail, we used a multivariate Gaussian proposal with zero mean
and a scaled diagonal covariance $\sigma I$, with $\sigma = 3^2$.

\begin{table}
\caption{Summary of \texttt{posteriordb} models}
\label{table:posteriordb}
    \centering
    \small
\begin{tabular}{ccc}
    \toprule
    Name & Dimension & Model description \\
    \midrule
    \texttt{kidscore} & 3 & linear model with a Cauchy noise prior \\
    \texttt{sesame} & 3 & linear model with uniform prior \\
    \texttt{gp\_regr} & 3 & Gaussian process regression with squared exponential kernel \\
    \texttt{garch11} & 4 & generalized autoregressive conditional heteroscedastic model \\
    \texttt{logearn} & 4 & log-log linear model with multiple predictors \\
    \texttt{arK-arK} & 7 & autoregressive model for time series \\
    \texttt{logmesquite} & 7 & multiple predictors log-log model  \\
    \texttt{8-schools} & 10 & non-centered hierarchical model for 8-schools \\
    \bottomrule
\end{tabular}
\end{table}

\begin{figure}
    \centering
    \includegraphics[scale=0.23]{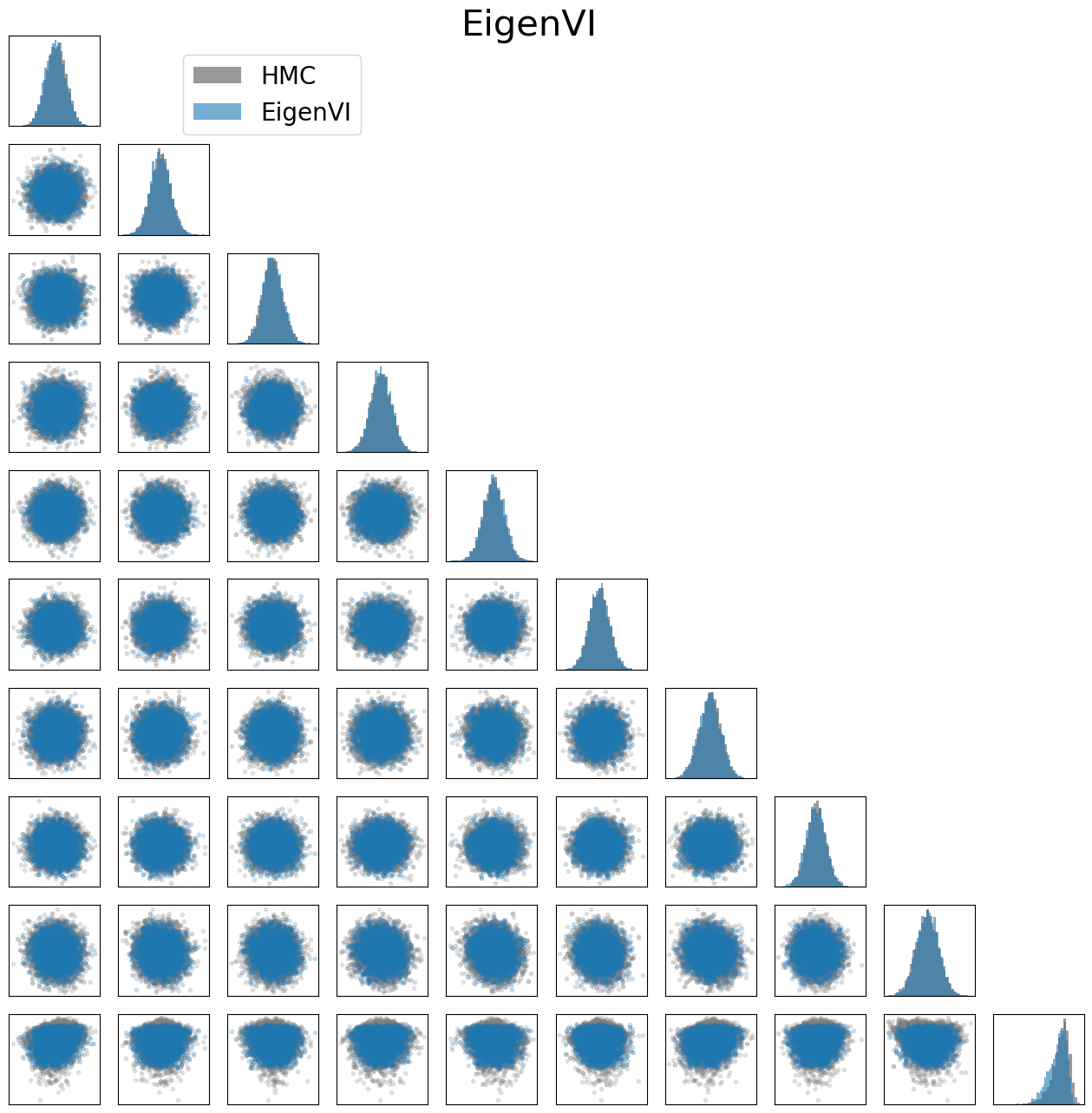}
    \\
    \includegraphics[scale=0.23]{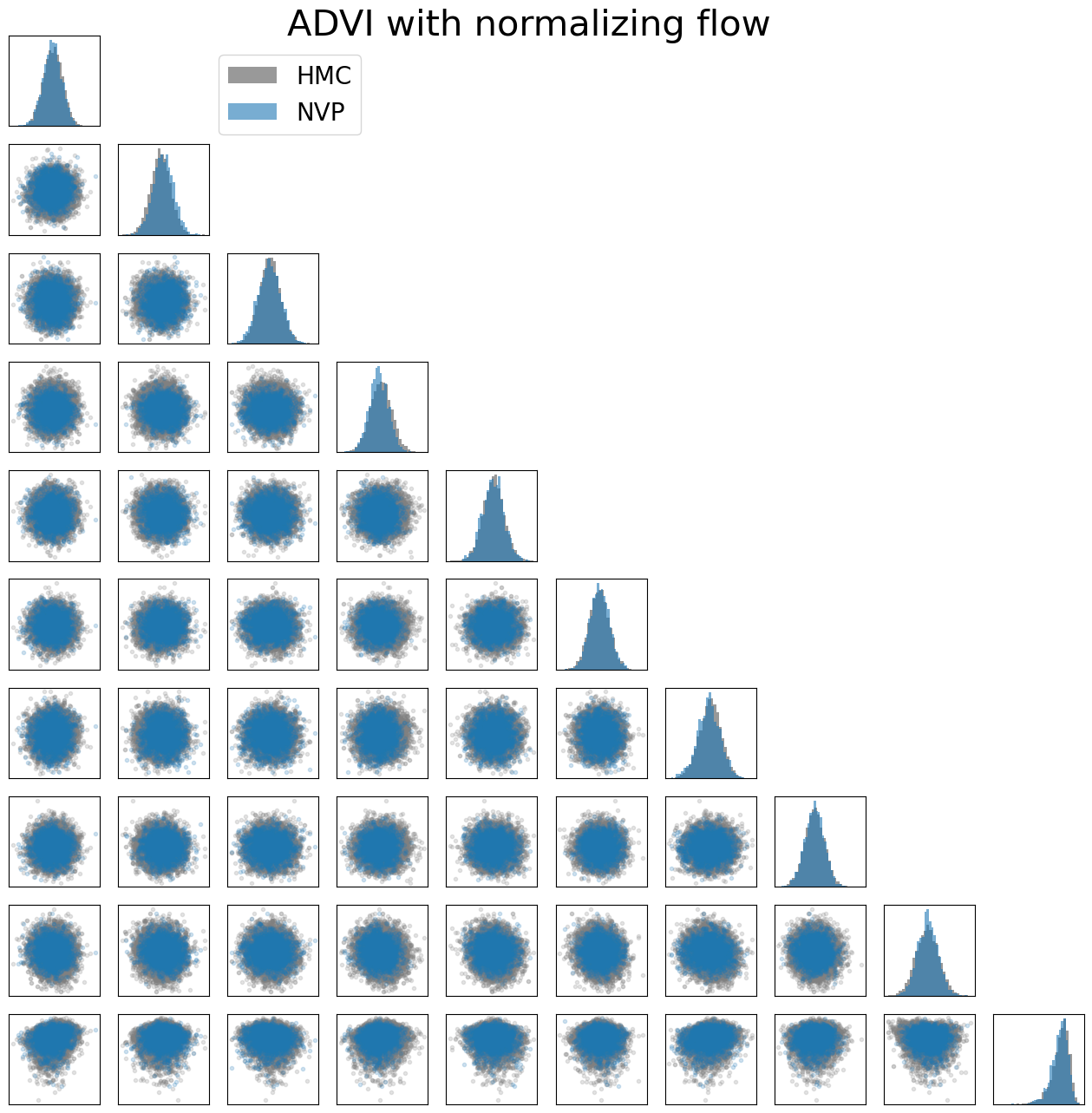}
    \\
    \includegraphics[scale=0.23]{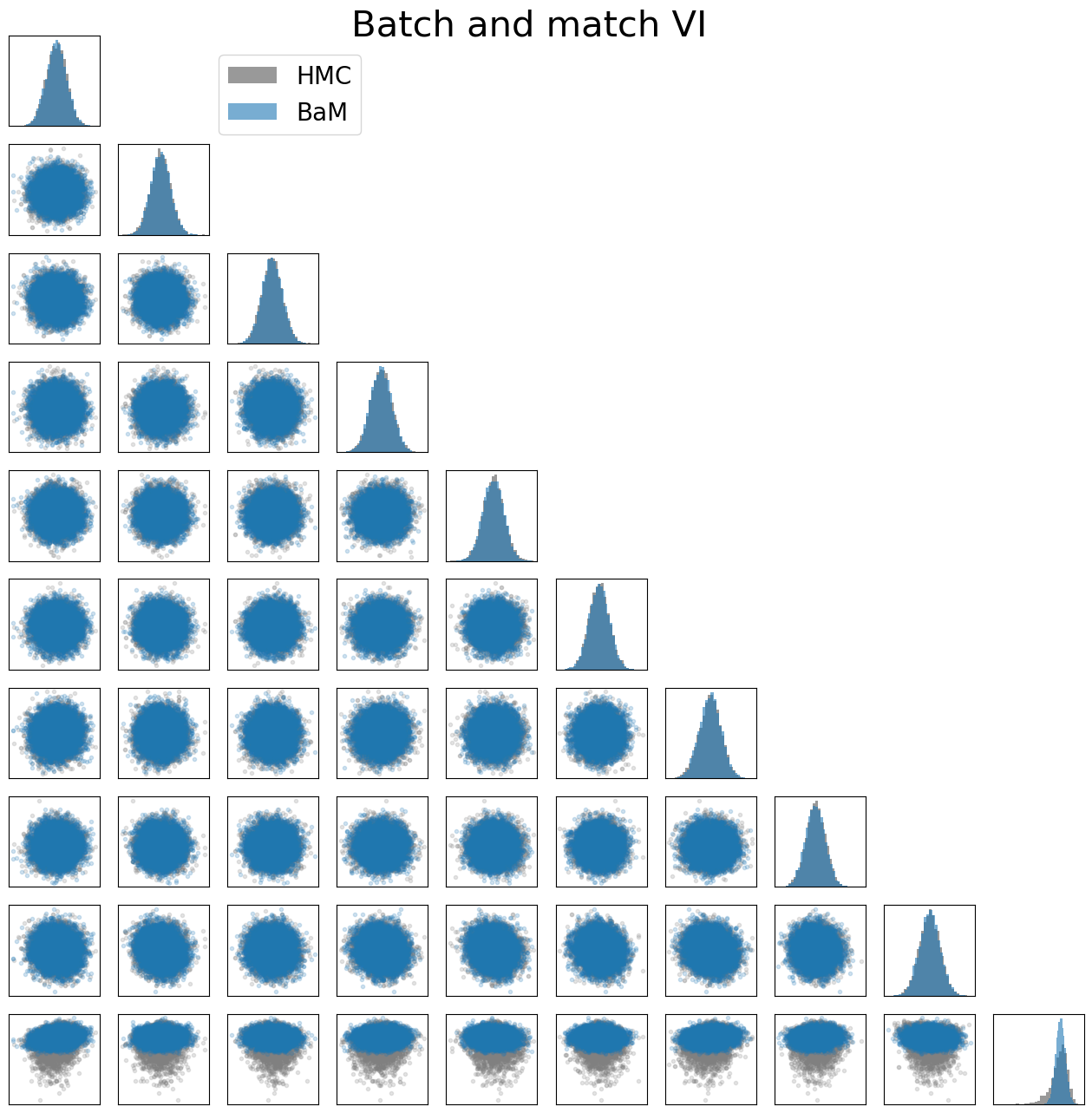}
    \caption{Comparison of EigenVI, normalizing flow,
    and Gaussian score-based BBVI methods on
\texttt{8schools}.
    }
\label{fig:8schools:corner}
\end{figure}

\begin{figure}
    \centering
    \includegraphics[scale=0.23]{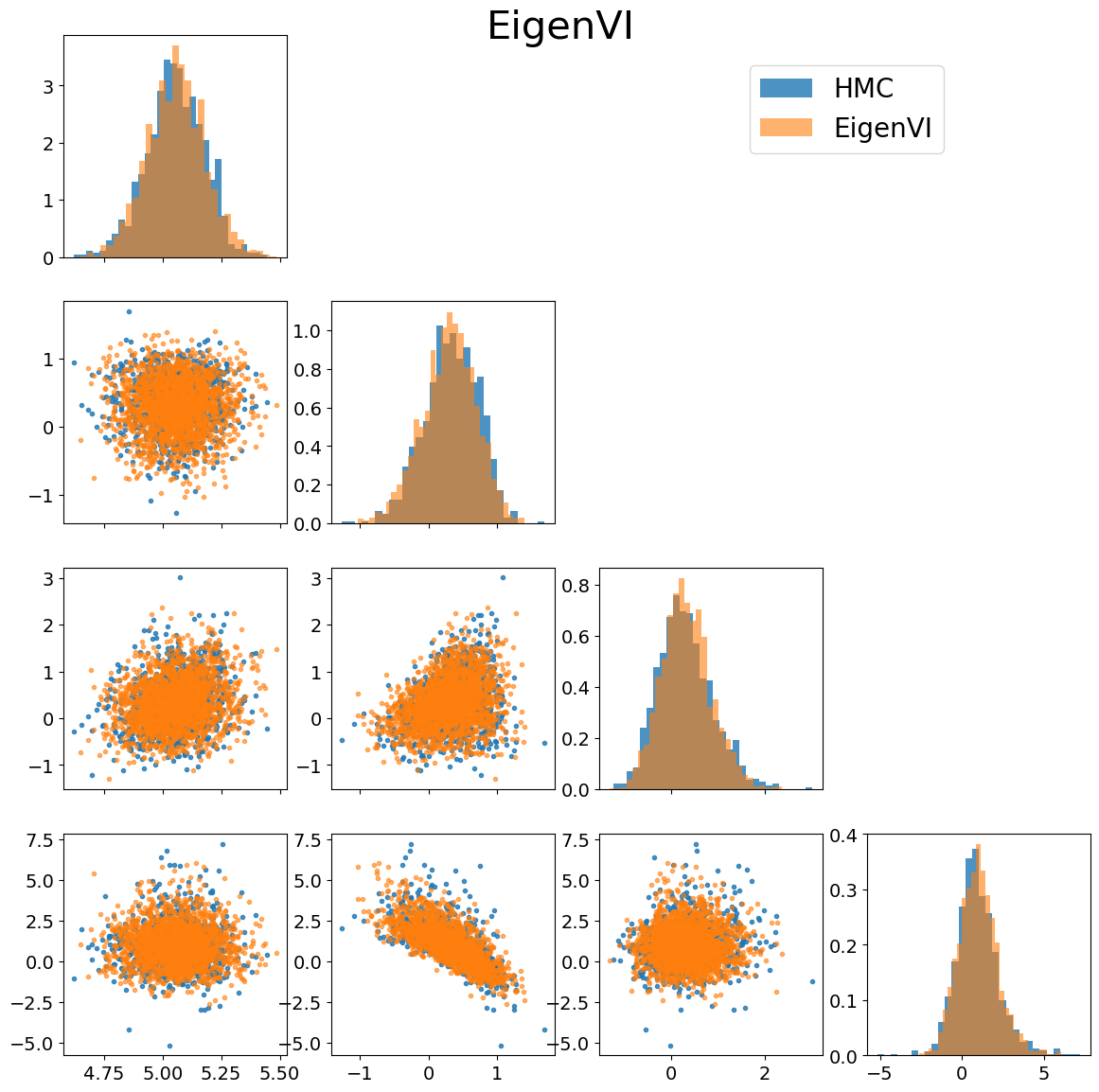}
    \\
    \includegraphics[scale=0.23]{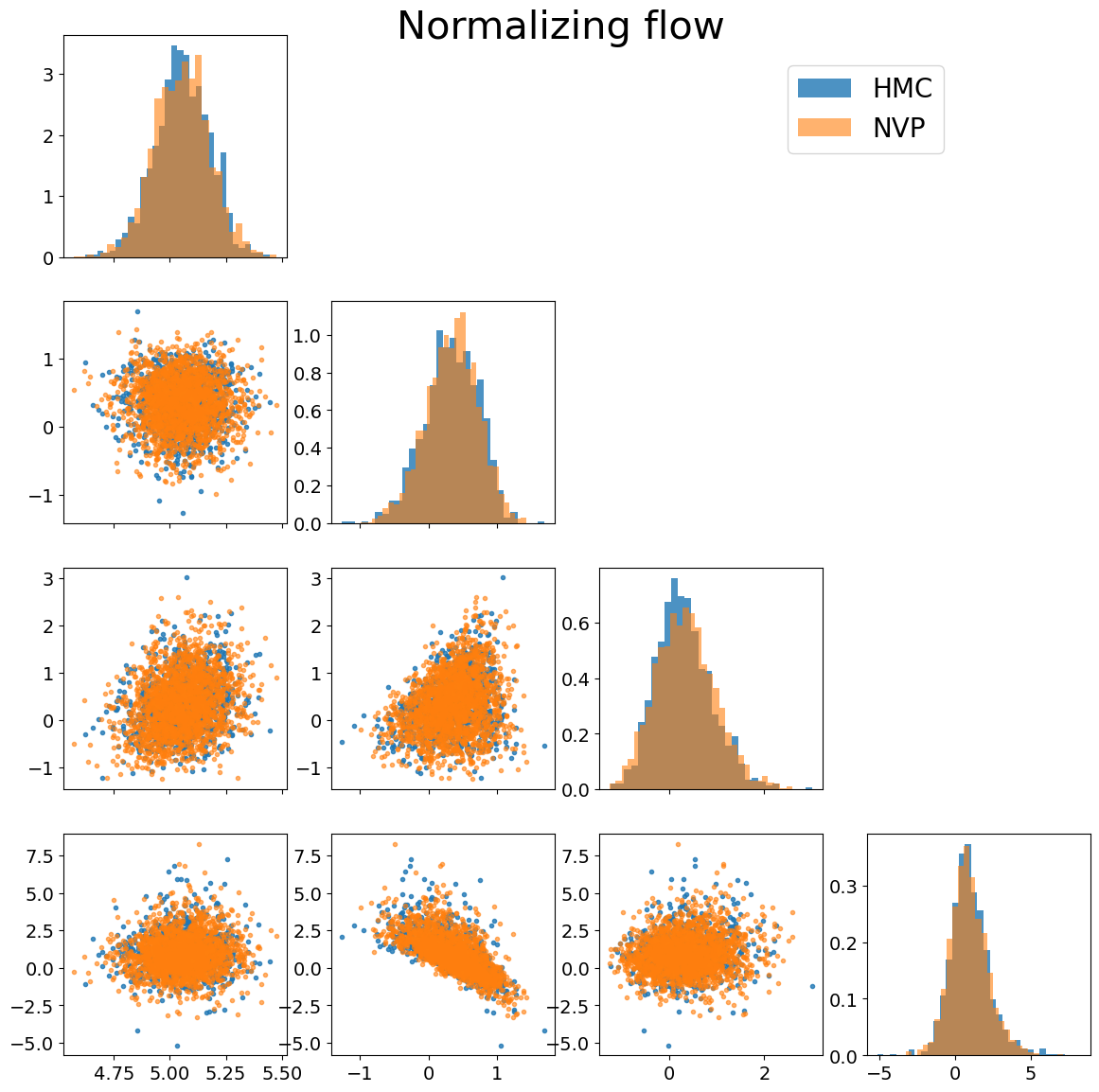}
    \\
    \includegraphics[scale=0.23]{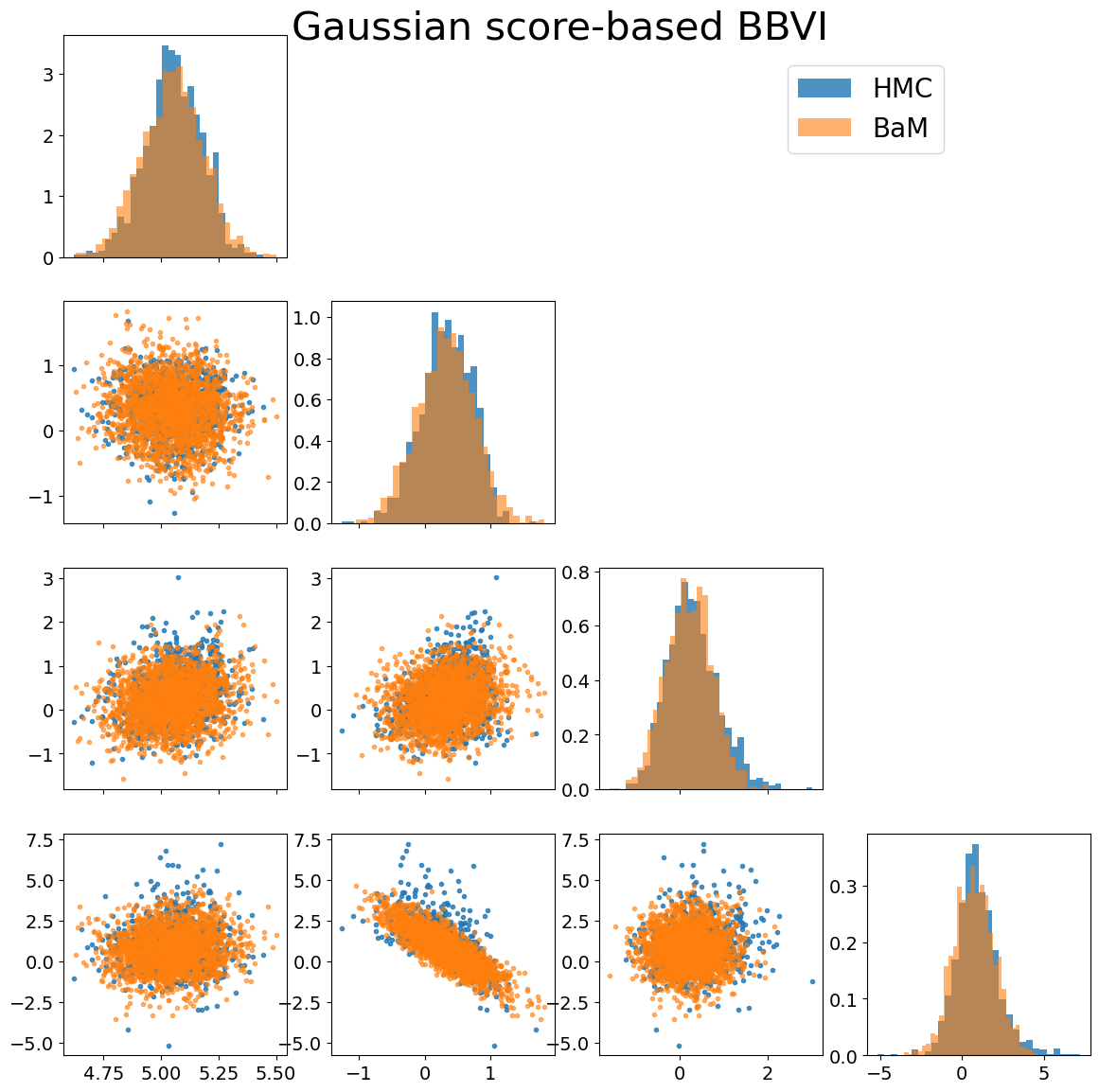}
    \caption{Comparison of EigenVI, normalizing flow,
    and Gaussian score-based BBVI methods on
\texttt{garch11}.
Note that the Gaussian approximation over/underestimates the tails,
while the more expressive families fit the tails better.
    }
\label{fig:garch11:corner}
\end{figure}

For the Gaussian score matching (GSM) method \citep{modi2023},
we chose a batch size of 16 for all experiments. We generally found the
results were not too sensitive in comparison to other batch sizes of 4,8, and 32.
For the batch and match (BaM) method \citep{cai2024}, we chose a batch size of 16.
The learning rate was fixed at
$\lambda_t = \tfrac{BD}{t+1}$, which was a recommended schedule for non-Gaussian targets.

For all ELBO optimization methods (full covariance Gaussian family and normalizing flow family),
we used Adam to optimize the ELBO.
We performed a grid search over the learning rate
$0.01, 0.02, 0.05, 0.1$ and batch size $B=4,8,16,32$.
For the normalizing flow model,
we used a real NVP \citep{dinh2016density}
with 8 layers and 32 neurons.
We found empirically that the computational of the scores was unreliable~\cite{kohler2021smooth,
Zeghal2022npe};
hence we do not show their Fisher divergence in \Cref{fig:posteriordb}.

In \Cref{fig:8schools:corner}
and
\Cref{fig:garch11:corner},
we show the corner plots that compare an EigenVI fit, a normalizing flow fit, and a Gaussian fit
(BaM).
In each plot, we plot the samples from the variational distribution against
samples from Hamiltonian Monte Carlo.
We observe that the two more expressive families EigenVI and the normalizing flow
are able to model the tails of the distribution better than the Gaussian fit.


\section{Broader impacts}

EigenVI adds to the literature on BBVI,
which has been an important line of work for developing automated, approximate
Bayesian inference methods.
In terms of positive societal impacts, Bayesian models are used
throughout the sciences and engineering, and advances in fast
and automated inference will aid in advances in these fields.
In terms of negative societal impacts, advances in BBVI could be
used to train generative models with malicious or unintended uses.


\end{document}